\documentclass[11pt, a4paper, reqno]{amsart}

\usepackage[top=2.9cm, bottom=2.9cm, left=2.5cm, right=2.5cm]{geometry}

\usepackage{fancyhdr}

\usepackage[utf8]{inputenc} 
\usepackage[T1]{fontenc}    
\usepackage{hyperref}       
\usepackage{url}            
\usepackage{booktabs}       
\usepackage{amsfonts}       
\usepackage{nicefrac}       
\usepackage{microtype}      
\usepackage{xcolor}         
\usepackage{amsmath, amssymb}
\usepackage{amsthm}
\usepackage{wrapfig}
\usepackage{float}
\usepackage{tikz}
\usepackage{tikzscale}
\usetikzlibrary{patterns,shapes.arrows}
\usepackage{algorithmic}
\usepackage[ruled]{algorithm2e}
\usepackage{microtype}
\usepackage{graphicx}
\usepackage{subfigure}
\usepackage{booktabs} 
\usepackage{hyperref}
\usepackage{framed}

\usepackage{mathtools}
\newtheorem{thm}{Theorem}[section]

\newtheorem{lem}[thm]{Lemma}
\newtheorem{prop}[thm]{Proposition}

\newtheorem{rem}[thm]{Remark}

\DeclareMathOperator{\Prob}{\mathbb{P}}
\DeclareMathOperator{\Exp}{\mathbb{E}}
\DeclareMathOperator{\Expemp}{\widehat{\mathbb{E}}}

\DeclareMathOperator{\erf}{erf}

\DeclareMathOperator{\srank}{\operatorname{srank}}
\DeclareMathOperator{\rank}{\operatorname{rank}}
\DeclareMathOperator{\vari}{\operatorname{Var}}
\DeclareMathOperator{\tr}{\operatorname{tr}}
\DeclareMathOperator{\cov}{\operatorname{Cov}}
\newcommand{\RR} {\mathbb R}

\newcommand{\T}{\intercal}
\renewcommand{\b}[1]{\mathbf{#1}}
\newcommand{\s}[1]{\boldsymbol{#1}}

\usepackage{xr}
\makeatletter
\newcommand*{\addFileDependency}[1]{
  \typeout{(#1)}
  \@addtofilelist{#1}
  \IfFileExists{#1}{}{\typeout{No file #1.}}
}
\makeatother

\newcommand{\GA}[1]{{\color{violet} {\bf (GA: #1)}}}

\usepackage{smartdiagram}

\title{On the Impact of Stable Ranks in Deep Nets}
\author[]{
    Bogdan Georgiev
    }\thanks{BG: Fraunhofer IAIS, ML2R, email: bogdan.m.georgiev@gmail.com}
    
\author[]{
    Lukas Franken
    }\thanks{LF: Fraunhofer IAIS, ML2R}

\author[]{
    Mayukh Mukherjee
    }\thanks{MM: IIT, Bombay}
    
\author[]{
    Georgios Arvanitidis
    }\thanks{GA: Max Planck Institute for Intelligent Systems, T\"ubingen}

\begin{document}

\maketitle

\begin{abstract}

A recent line of work \cite{arora:icml:2018, Bartlett-NIPS2017, neyshabur:neurips:2017, sanyal:iclr:2020} has established intriguing connections between the generalization/compression properties of a deep neural network (DNN) model and the so-called layer weights' stable ranks. Intuitively, the latter are indicators of the effective number of parameters in the net. In this work, we address some natural questions regarding the space of DNNs conditioned on the layers' stable rank, where we study feed-forward dynamics, initialization, training and expressivity. To this end, we first propose a random DNN model with a new sampling scheme based on stable rank. Then, we show how feed-forward maps are affected by the constraint and how training evolves in the overparametrized regime (via Neural Tangent Kernels). Our results imply that stable ranks appear layerwise essentially as linear factors whose effect accumulates exponentially depthwise. Moreover, we provide empirical analysis suggesting that stable rank initialization alone can lead to convergence speed ups.



\end{abstract}

\section{Introduction}
\label{sec:introduction}

\textbf{Stable Ranks in Neural Networks}

\begin{figure}
\label{fig:three_spheres}
  \begin{center}
    \includegraphics[width=0.48\textwidth]{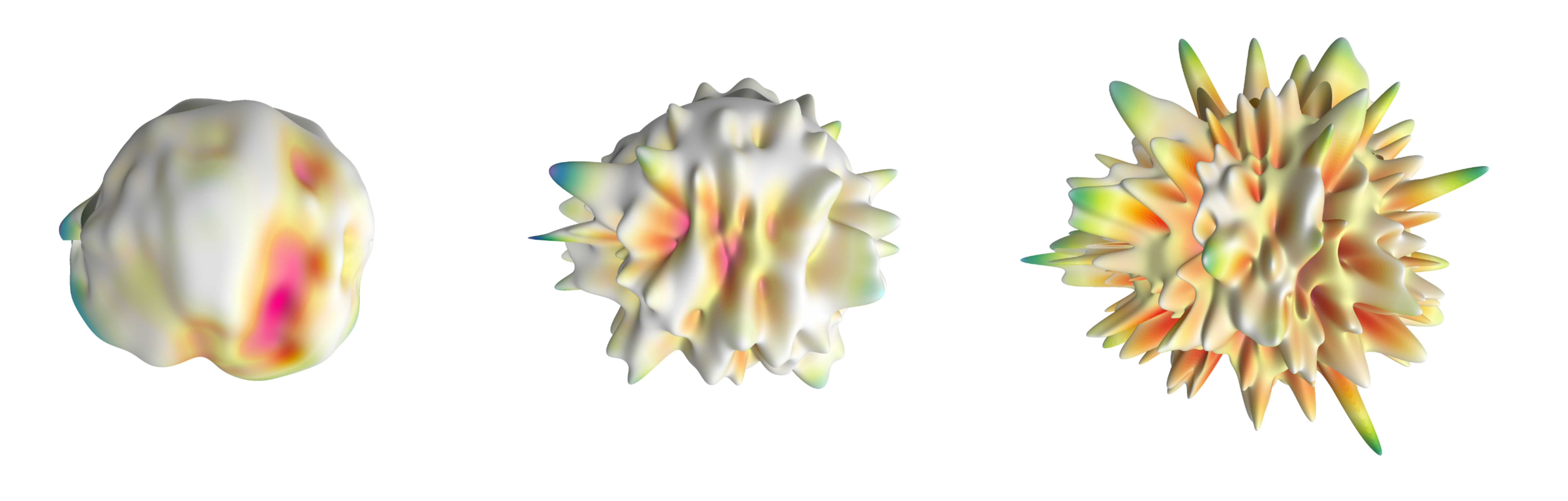}
  \end{center}
  \caption{The geometric features and (PAC-based) generalization bounds of DNN's hidden representations are controlled by the layer weights' stable rank via \textit{exponential-to-linear} trade-off. Left to right: the output of a multi-layer perceptron whose layer weights are of increasingly large stable ranks.}
\end{figure}

Obtaining good generalization properties is a central task in current machine learning research as well as practical work. Motivated by classical VC theory \cite{Vapnik1999}, a recent line of work \cite{arora:icml:2018, neyshabur:neurips:2017, Bartlett-NIPS2017} obtained DNN generalization-error estimates in terms of the so-called stable ranks of the weight matrices (i.e. the ratio of the matrix's Frobenius and spectral norms; formal definition is given below). Intuitively, the stable rank is a continuous counterpart of the rank of a matrix and represents the effective ``true'' number of parameters, or the ``effective'' size of a layer in the DNN. Hence, bounds on the stable rank yield control on the complexity of the weight matrices and, ultimately, on the flexibility of the hidden representations. Recently, stable rank-based regularization was studied \cite{sanyal:iclr:2020}, and furthermore, \cite{sanyal:iclr:2020, miyato2018spectral} empirically demonstrated that carefully crafted regularizations based on spectral norms and stable ranks can lead to impressive performance improvements of DNN models in terms of scaling and generalization (in particular, for Generative Adversarial Networks \cite{Goodfellow2020}).

A major question that was implicitly addressed in the above body of literature asks about the DNN's capabilities whenever stable ranks are constrained: 

\textbf{Question 1:} \textit{How can one quantify the trade-off between lower stable ranks, generalization bounds and the network's expressivity? In particular, how "expressive" can the space of DNNs conditioned on their layers' stable ranks be?}

A missing piece of analysis in this regard goes along the net's geometric and feed-forward map properties. The notion of DNN expressivity has been central to a substantial body of ML literature (e.g. \cite{raghu:icml:2017, poole:neurips:2016} and the references therein). By expressivity, one usually refers to the network's ability to adapt to a multitude of scenarios and data distributions, still providing meaningful and useful data representations w.r.t. the present task (e.g. classification, clustering, etc). To this end, a highly-expressive neural net should geometrically be capable of shaping and transforming the data in a sufficiently sophisticated way. Hence, our goal in this regard will be to analyze how basic geometric notions such as lengths of curves, distances and curvature change as one applies the feed-forward map of the neural network.

\textbf{Question 2:} \textit{Is the space of DNNs conditioned on stable ranks useful for initialization? How is this reflected in the training?}

Previous work in this regard established mostly empirical results on stable rank regularized training \cite{sanyal:iclr:2020}. However, the theoretical training framework appears to a large extent unexplored. Moreover, interesting related questions if (and why) stable rank initialization alone is beneficial remain open.

\textbf{In this work} we discuss Questions 1, 2 in the regime of large-width DNNs. Regarding Question 1: 
\begin{itemize}
    \item We propose and study a random DNN model constrained on layers' stable rank (Section \ref{sec:sampling}).
    \item After obtaining suitable moment bounds for our non-standard layer sampling, we theoretically study the behavior in the large width limit. Here, we show how one can make use of both simultaneous limit techniques \cite{Yang_2019, matthews:iclr:2018} as well as "classical" sequential (GP-based) limits \cite{jacot:neurips:2018, poole:neurips:2016, raghu:icml:2017, neal:report:1994}. Regarding the latter and somewhat surprisingly, our non-standard matrix sampling still yields normal distributions via an application of a non-trivial Central Limit Theorem for the Orthogonal group $O(n)$ as $n \rightarrow \infty$ \cite{aristotile03}.
    \item Finally, we obtain control on the DNN's feed-forward dynamics and see that stable ranks act as linear factors layerwise (Theorem \ref{thm:gp-ff}). Using arguments in \cite{poole:neurips:2016, raghu:icml:2017} one sees that within our DNN model small stable ranks (bounded above by $s$) lead to an \textit{exponential decrease} of geometric expressivity (regarding the net's depth $L$) in terms of curvature and curve lengths, i.e. these geometric quantities are bounded by $\sim s^L$ (Section \ref{sec:srank-expressivity}). At the same time the model-capacitory contribution in the well-known theoretical PAC-based generalization gaps \cite{neyshabur:iclr:2018, arora:icml:2018, sanyal:iclr:2020} are \textit{improved linearly} in terms of the net's stable rank, i.e. as $\sim L s$.
\end{itemize}

Regarding Question 2:

\begin{itemize}
    \item We analyze training of stable rank constrained nets via Neural Tangent Kernels (Section \ref{sec:ntk-evolution}). Similar to the feed-forward dynamics, we see that  evolution kernels scale layerwise w.r.t.  stable ranks.
    \item Further, we provide empirical analysis and theoretical motivation as to why stable rank initialization alone could be useful for improving training convergence (Section \ref{sec:experiments}).
\end{itemize}

\textbf{Further motivation and some open questions}

In the present paper, the random model methodology is inspired by works such as \cite{poole:neurips:2016, raghu:icml:2017} - namely, our considerations are mostly taking place within a suitable random DNN model where the layers are stable rank constrained. Furthermore, as in \cite{poole:neurips:2016, raghu:icml:2017} some of our analysis (and proposed random DNN model) is not making use of data characteristics or training-algorithm information. The guiding idea here is that one attempts to understand generic DNN behavior under stable-rank (capacity) constraints. Moreover, as a first step we implicitly interpret these constraints as well-controlling the PAC-based generalization bounds (Section \ref{sec:background} below). Of course, the full generalization bound also includes a post-training loss term (which in practice is often comparable or negligible in comparison to the other model-capacitory terms) - in this regard one interpretation of our random DNN model is as follows: one trains a black-box DNN without direct access to the layer weights, but after training has information only on stable-ranks and train loss (thus controlling completely the theoretical generalization bound); hence, "opening the black-box" could be interpreted in a simplistic way as drawing a DNN from the proposed scheme. Indeed, a natural follow-up would be to improve the random DNN model by explicitly using training or data information. Below we observe how data sets might exhibit ``natural'' stable ranks - being able to estimate those in advance is an intriguing open problem. In particular, this might require one to understand and extract insights on how the layers' stable ranks are related with data Hessian and NTK evolution in the spirit of \cite{Arora_fine_grain_ICML}.

Another interesting open question addresses the negative correlation between generalization and some PAC-based estimates established empirically for unconstrained/unregularized random DNNs \cite{Jiang*2020Fantastic}. It is argued in \cite{sanyal:iclr:2020} that this observation becomes invalid whenever stable rank constraints/regularization are enforced within this random DNN model. Here, we also provide some empirical analysis in support of the latter. However, the intriguing question how and in which regime the negative correlation holds remains open.

\section{Background}
\label{sec:background}
We define a \textit{fully connected feed-forward DNN} with Lipschitz activation $\varphi$, and $L$ layers of width $N_0, \dots, N_L$ as a mapping $f_\theta:\mathbb{R}^{N_0} \rightarrow \mathbb{R}^{N_L}$ ($\b{\theta}$ being the set of parameters) with the following structure. For $l = 1, \dots, L$ let $\b{W}^l$ be the $N_l \times N_{l - 1}$ matrix in the $l$-th layer, and $\b{b}^l$ be the corresponding bias. We let $\tilde{\s{\alpha}}^l$ be the output (pre-activation) from the $l$-th layer, and $\s{\alpha}^l$ be the post-activation. Thus, for an input vector $\s{x} \in \mathbb{R}^{N_0}$ the feed-forward mappings are given by:
\begin{align}\label{eq:Poole-ff-dynamics}
    \tilde{\s{\alpha}}^{l}(\s{x}) & = \gamma^l\b{W}^l \s{\alpha}^{l-1}(\s{x}) + \b{b}^l, \quad
    \s{\alpha}^{l}(\s{x}) = \varphi (\tilde{\s{\alpha}}^l(\s{x}))
\end{align}
where we set $\s{\alpha}^{0}(\s{x}) = \s{x}, f(\s{x}) =  \tilde{\s{\alpha}}^{L}(\s{x})$ and the nonlinear activation function $\varphi(\cdot)$ is applied coordinate-wise. $\gamma^l$ is a normalization factor, and the common choice for $\gamma^l$ in ML literature is $1$. However, to ensure stability at the large width asymptotics, some results require a normalization factor which compensates for the growing width (e.g., in  \cite{jacot:neurips:2018, Yang_2019}, the authors have used $\gamma^l = \frac{1}{\sqrt{N_{l- 1}}}$). Note that in this context, the parameter set $\s{\theta}$ consists of the layer widths $\b{W}^l$ and biases $\b{b}^l$. Under the standard random Gaussian weights assumption the entries of each weight matrix $\b{W}^l \in \mathbb{R}^{N_l\times N_{l-1}}$ at the layer $l$ are sampled independently from a normal distribution $\mathcal{N}(0, \sigma_w^2 / N_{l-1})$, and similarly, the bias entries are initially sampled from $\mathcal{N}(0, \sigma_b^2)$ where $\sigma_w, \sigma_b$ are some fixed positive numbers.

For a matrix $\b{A}\in \RR^{D_1 \times D_2}$ we denote by $\|\b{A}\|_2$ the \textbf{spectral norm} of the operator, i.e. $\sup_{\|\s{x}\| = 1} \|\b{A}\s{x}\|_2$; $\|\b{A}\|_2$ is the largest singular value of the linear operator $\b{A}$. The \textbf{stable rank} of a matrix $\b{A}\in \RR^{D_1 \times D_2}$ is defined as
\begin{equation} \label{def:srank}
    \srank(\b{A}) := \frac{\|\b{A}\|_F^2}{ \|\b{A}\|_2^2} = \frac{\sum_{i=1}^{\rank(\b{A})} \sigma_i^2}{\sigma_1^2},
\end{equation}
where $\|\cdot\|_F$ denotes the Frobenius norm and $\sigma_1 \geq \dots \geq \sigma_{n}$ denote the singular values of $\b{A}$, while $\rank(\b{A})\leq\min(D_1, D_2)$. As mentioned before, the stable rank is intuitively understood as a continuous proxy for $\rank(\b{A})$ and is known to be more stable against small perturbation \cite{arora:icml:2018, sanyal:iclr:2020, neyshabur:neurips:2017}.

The relation between stable ranks and generalization 
has been intensively studied over the last years \cite{sanyal:iclr:2020}. For a DNN consisting of $L$ layers as above, recent works obtain generalization errors, roughly speaking, in terms of the expression $\mathcal{O}\left(\prod_{l=1}^L \|\b{W}^l\|^2_2 \sum_{l=1}^L \srank(\b{W}^l)\right)$, see 
\cite{neyshabur:iclr:2018, arora:icml:2018}. A related stronger compression-based estimate in terms of so-called \emph{noise cushions} is obtained in \cite{arora:icml:2018}. Clearly, the generalization error is influenced by the spectral norms, as well as the stable ranks of the layers. Intuitively, small stable rank and spectral norm implies a smoother model, while for high values the learning model can overfit the training data.


\section{Random DNNs with given stable rank and spectral norm} 
\label{sec:sampling}

To obtain a basic intuition how stable ranks might vary, we first evaluate the stable rank behaviour of the standard Gaussian model for random DNNs as defined in Sec.~\ref{sec:background}.

\begin{prop}[Supplementary, Section \ref{sec:Sampling_properties}] \label{prop:srank-mean-field-nn}
    Let us suppose that the DNN's weights are i.i.d. Gaussian and the widths $N_l, N_{l-1}$ are increasingly large, but the ratio  $N_{l-1} / N_{l} = \alpha $ is kept fixed for some $\alpha \in (0, 1]$. Then, for the weights $\b{W}^l$ at layer $l$ of a random DNN as proposed in Sec.~\ref{sec:background}, we have
    \begin{equation}
        \frac{\srank(\b{W}^l)}{N_{l}} \rightarrow \frac{\alpha }{(1 + \sqrt{\alpha})^2},
    \end{equation}
    almost surely as  $N_l, N_{l-1} \rightarrow \infty$. In particular, for a constant large width $N_l$ across layers $l = 1, \dots, L$, the stable rank $\srank(\b{W}^l)$ is essentially $N_l / 4$.
\end{prop}

In other words, this statements sheds light on what implicit stable ranks some previous expressivity studies (examining i.i.d. Gaussian weights) have dealt with - e.g. in \cite{poole:neurips:2016} the authors have empirically examined DNNs with $N_l = 1000$, and hence worked with matrices of approximate stable rank $250$. As next step, it is natural to inquire about the behaviour of a DNN (expressivity and geometry propagation) provided the stable ranks are controlled explicitly. To this end, one would need to design a model where the weights $\b{W}^l$ are suitably sampled according to their stable rank. In the following we will define our weight sampling methods and obtain results for the feed-forward dynamics. To accomplish this, we first need to be able to sample weight matrices $\b{W}^l$ with respect to a fixed target stable rank $r_t \geq 1$ and spectral norm $s_t$. We propose a sampling strategy based on the singular value decomposition

\begin{equation}
\label{eq:stable_rank_spectral_norm_weights}
    \b{W}^l = \b{U} \boldsymbol{\Sigma} \b{V}^\T, \quad \text{where }
        \b{U} \sim  \mathcal{U}\left(O(N_l)\right),
        \b{V} \sim \mathcal{U}\left( O(N_{l-1}) \right),
        \boldsymbol{\Sigma} \sim \mathcal{S}_{r_t, s_t}
\end{equation}

Here $\mathcal{U}\left(O(N_l)\right)$ denotes the uniform distribution w.r.t. Haar measure on the orthogonal group of dimension $N_l$ and $\mathcal{S}_{r_t, s_t}$ denotes a uniform distribution in spectral space over diagonal matrices with given stable rank and spectral norm. Indeed, stable ranks and spectral norms are invariant under orthogonal transformations, so the resulting matrix $\b{W}^l$ will posses the required properties.

\subsection{Uniform sampling over matrices with given stable rank and spectral norm} \label{subsec:uniform-lebesgue-sampling}

\begin{wrapfigure}[22]{R}{0.5\textwidth}
    \begin{algorithm}[H]
    \SetCustomAlgoRuledWidth{0.45\textwidth} 
    \caption{Sampling diagonal matrix $\boldsymbol{\Sigma}$ for Eq.~\ref{eq:stable_rank_spectral_norm_weights}}
    \label{alg:naive_sampling}  
        
        \SetKwData{Left}{left}\SetKwData{This}{this}\SetKwData{Up}{up}\SetKwFunction{Union}{Union}\SetKwFunction{FindCompress}{FindCompress}\SetKwInOut{Input}{input}\SetKwInOut{Output}{output}\SetKwInOut{Ensure}{ensure}
        
         \Input{Stable rank $r_t$, spectral norm $s_t$, sizes  $N_l,~N_{l-1}$}
         	
         \Ensure{Diagonal matrix $\boldsymbol{\Sigma}\in\RR^{N_l \times N_{l-1}}$ with provided $r_t$ and $s_t$.}
         Set $s_1 := 1$ and $m := \min(N_l, N_{l-1})$ \label{alg:naive-sampling:step_1} 
         	
         \For{$i = 2, \dots, m$}{Set $s_i := |x_i|$, with i.i.d. $x_i \sim \mathcal{N}(0, 1) $ \label{alg:naive_sampling:step_2}}
         Define $r := \frac{\sqrt{r_t - 1}}{ \sqrt{\sum_{i=2}^{m} s_i^2}} $ \label{alg:naive_sampling:step_3} \\
         Rescale $s_i := r * s_i, i = 2, \dots, m$ \label{alg:naive_sampling:step_4}
         	
         \uIf{$s_1 \geq \max_{i=2, \dots, m} s_i$}{Return $\boldsymbol{\Sigma} := \text{diag}(s_t * \{s_i\}_{i=1}^m) \in \mathbb{R}^{N_l \times N_{l-1}}$ \label{alg:naive_sampling:step_5}}
         \Else{Go to Step~2}{}
         	
    \end{algorithm}
\end{wrapfigure}

A straightforward way to uniformly sample matrices as in  Eq.~\ref{eq:stable_rank_spectral_norm_weights} is via rejection sampling, outlined in Algorithm~\ref{alg:naive_sampling}. In addition, one can come up with further (reasonable) sampling methods - e.g. in the Supplementary, Subsection \ref{subsec:further-sampling} we discuss a method based on uniform sampling over a cube in spectral space.


\begin{prop}[Supplementary, Subsection \ref{subsec:correctness}]\label{prop:samp_prod}
    Algorithm~\ref{alg:naive_sampling} produces a sample that follows uniform distribution over all diagonal matrices with given spectral norm $s_t = s_{11}$ and stable rank $r_t$.
\end{prop}
As usual, the rejection sampling in higher dimensions and $r_t $ close to $m$ is to be taken with care. In order to guarantee that the algorithm ends faster one could use various further techniques from MCMC or instead of starting with uniform samples on the sphere (Steps~3-5), one could sample points from a spherical cap of appropriate radius. 
\begin{prop} [Sampling efficiency, Supplementary, Subsection \ref{subsec:correctness}] \label{prop:gaussian-acceptance-rate}
    Let $\eta$ be an arbitrary positive real number that is at least $\sqrt{(r_t-1) / (m-1)}$. Then, the above algorithm accepts a sample with probability at least
    \begin{align}
        \Prob\left[ \text{Accept } \{ s_i \} \right] \geq 1 &- \left( \frac{r_t - 1}{\eta^2 (m-1)} e^{1 - \frac{r_t - 1}{\eta^2(m-1)}} \right)^{\frac{m-1}{2}} - (m-1)\left( 1 - 2 \erf \left( \frac{1}{\eta} \right) \right).
    \end{align}
    In particular, if the dimension $m$ is sufficiently large and 
    $r_t / m$ is sufficiently small, one can select $\eta$ very close to $0$ and conclude that the above algorithm accepts a sample with large probability.
\end{prop}
\section{Large width asymptotics of the random stable rank DNN model}
\label{sec:kernel_methods}

We now study the behaviour of feed-forward neural networks whose layer weights are initialized by the proposed algorithm. We emphasize that, in contrast to the Gaussian initialization, our weight sampling produces non-independent and non-Gaussian entries. Thus, the analysis of the net's mappings requires further tools such as specialized law-of-large-numbers and central limit theorems. 

{\bf Sequential vs simultaneous limits} While investigating large width DNNs, we recall that there are by now two popular ways in the literature to interpret the large width regime: (a) the so-called ``sequential limits'' where the large width limits are taken iteratively through the layers, as popularized by, e.g., \cite{neal:report:1994, jacot:neurips:2018} "geometrically" in \cite{poole:neurips:2016, raghu:icml:2017}; (b) the process of ``simultaneous limits'', where all layer widths are allowed to go to infinity at the same time \cite{Yang_2019, matthews:iclr:2018}. In particular, the simultaneous limit techniques provide a general framework where one does not require independence of the individual weights, but rather only that the individual neurons are exchangeable within the
same layer, which is also true for our initialization. In the statements following henceforth, we will use the nomenclature ``in the large width regime'' to encompass both the above. It is well-known that i.i.d. Gaussian weight sampling yields DNNs that behave like Gaussian processes provided the layer widths are sufficiently large \cite{Neal1996, lee:iclr:2018, jacot:neurips:2018, Yang_2019}. However, it is somewhat surprising that even with our non-i.i.d. we are able to recover a similar phenomenon, which seems to give a strong indication that the sampling described by Algorithm 1 is natural and useful for further purposes (e.g. Bayesian inference and kernel analysis \cite{Shawe-Taylor2004}). Moreover, provided the spectral norms are held fixed, the feed-forward dynamics is step-wise linearly influenced by the respective stable ranks - lower stable ranks would yield a decrease in lengths and curvatures as we see below. Our main result in this Section prescribes the feed-forward dynamics at initialization as a Gaussian Process (both in the sequential and simultaneous limit regimes):

\begin{thm}[GP Behavior, Supplementary, Section \ref{sec:proof_large_width_single_layer}] \label{thm:gp-ff}
Let us suppose that $\b{W}^l$ are sampled as in Algorithm \ref{alg:naive_sampling} with given target spectral norm $s_l$ and stable rank $r_l$, $l = 1, \dots, L$. At initialization, in the sequential limit $N_1, \dots, N_{l - 1} \nearrow \infty$ the output functions $\tilde{\alpha}_k^l(x, \s{\theta})$, and in the simultaneous limit the final output function $f_{\theta, k}(x)$ 
tend to centered Gaussian processes of covariance $\s{\Sigma}^l$ (respectively $\s{\Sigma}^L$), where $\s{\Sigma}^L$ is defined in the following recursive way:
\begin{align}
    \s{\Sigma}^1(\s{x}, \s{x}') & = 
    \frac{s_1^2 r_1}{N_0N_1}\langle \s{x}, \s{x}'\rangle + \sigma_b^2,\\
    \s{\Sigma}^{l}(\s{x}, \s{x}') & = 
    \frac{s_l^2 r_l}{N_{l-1}N_l} \Exp_f \left( \varphi(f(\s{x})) \varphi (f(\s{x}')) \right) + \sigma_b^2,
\end{align}
where $\gamma^l = \frac{1}{\sqrt{N_{l-1}}}$ for $l = 1, \dots, L $ and  the expectation in the last step is taken with respect to a centered Gaussian process $f$ of covariance $\s{\Sigma}^{l-1}$.
\end{thm}

\begin{figure}
    \centering
    \includegraphics[width=1\linewidth]{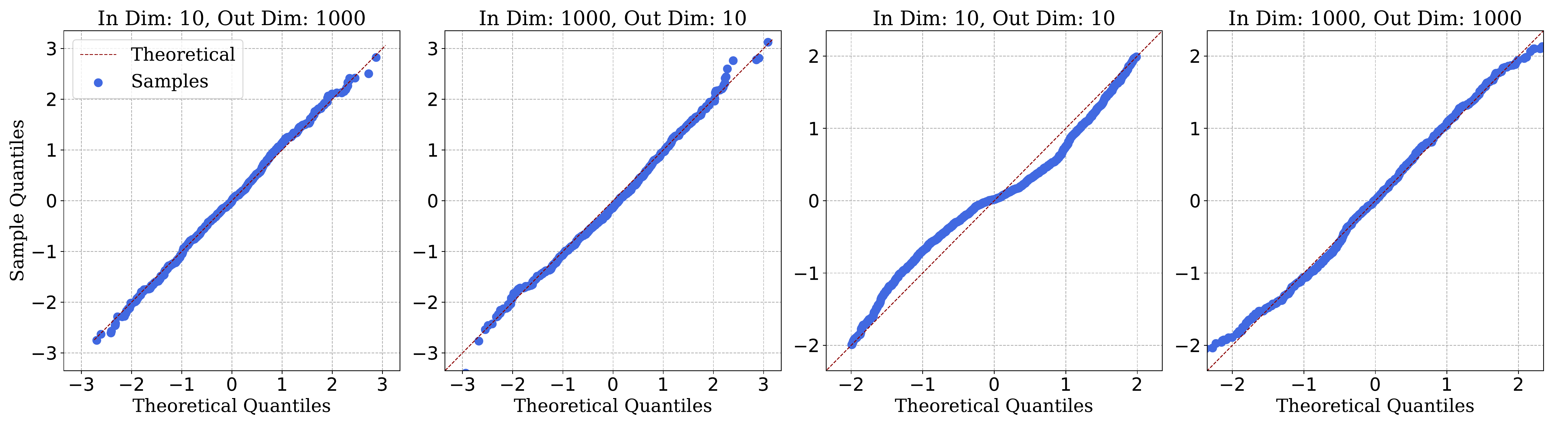}
    \caption{QQ-plots illustrating the convergence to a Gaussian distribution for a single neural layer in different input/output regimes as described in Proposition \ref{lem:large_width_behavior}. In each subplot a random input point $\s{x}$ is fixed and the Gaussian distribution is rescaled in accordance with 
    Proposition \ref{lem:large_width_behavior}.}
    \label{fig:large-witdth-asymptotics}
\end{figure}

The proof of Theorem \ref{thm:gp-ff} is rather technical, and we indicate the main steps below. In order to analyze the network's feed-forward map one first needs to evaluate the covariance of the weight matrices in our sampling scheme. To this end one uses results from representation theory and Weingarten functions \cite{aristotile03}:

\begin{prop}[Supplementary, Section \ref{sec:proof_large_width_single_layer}] \label{prop:srank-covariance}
    According to the sampling above, let $\b{W}^l = \b{U} \boldsymbol{\Sigma} \b{V}^\T$ where $\b{U},~\b{V}$ are orthogonal matrices independently uniformly sampled w.r.t. Haar measure and where $\boldsymbol{\Sigma} \sim \mathcal{S}_{r_t, s_t}$. Then,
    \begin{equation}\label{eq:layer_expectation_1}
        \Exp \left[W^l_{ij} W^l_{kl} \right] = \delta_{ik} \delta_{jl} \frac{s_t^2 r_t}{N_l N_{l-1}}.
    \end{equation}
\end{prop}
Second, the main technical step is the single-layer dynamics, summarized in the following:
\begin{prop}[Large-width behavior of a single layer, Supplementary, Section \ref{sec:proof_large_width_single_layer}]\label{lem:large_width_behavior}
Let us suppose that $\b{W} = \b{USV^T} \in \mathbb{R}^{N_0 \times N_1}$ is sampled as in Section \ref{sec:sampling}. Then, provided the bias' variance $\sigma_b$ is chosen sufficiently small w.r.t. widths $N_0, N_1$, we have the following convergence properties:
    \begin{enumerate}
        \item Conditioning on $\b{U}, \b{S}$, as $N_0 \rightarrow \infty$ the 
        outputs 
        $f_{\s{\theta}, k}(\s{x})$ in distribution approach a centered Gaussian process (indexed by $\s{x}$) with covariance $\s{\Sigma}(\s{x}, \s{x}') = \frac{\langle \s{x}, \s{x}' \rangle}{N_0} \sum_a (U_{ka} S_{aa})^2 + \sigma_b^2$.
        \item Conditioning on $\b{V}, \b{S}$, as $N_1 \rightarrow \infty$ the 
        outputs 
        $f_{\s{\theta}, k}(\s{x})$ in distribution approach a centered Gaussian process (indexed by $\s{x}$) with covariance $\s{\Sigma}(\s{x}, \s{x}') = \sum_a \frac{S_{aa}^2}{N_1} \sum_j (V_{aj} x_j x_j')^2 + \sigma_b^2$.
        \item If 
        one lets $N_0, N_1 \rightarrow \infty$ 
        simultaneously, 
        then the conditioning on $\b{U}, \b{S}, \b{V}$ does not play a role and the outputs $
        f_{\s{\theta}, k}(\s{x}) $ approach a centered Gaussian process with covariance $\s{\Sigma}(\s{x}, \s{x}') = \langle \s{x}, \s{x}' \rangle \frac{s^2 r}{N_0 N_1} +  \sigma_b^2$.
    \end{enumerate}
    Numerical illustration of the convergence is given in Fig. \ref{fig:large-witdth-asymptotics}.
    \end{prop}
The above proposition sets up the base case for the induction for multilayer DNNs. In particular, Statement 3. above allows us to carry through the induction for the simultaneous limits regime.
    
\section{Effects on geometry and expressivity}
\label{sec:srank-expressivity}
In \cite{poole:neurips:2016} and \cite{raghu:icml:2017} the authors start with a random DNN model with i.i.d. Gaussian weights. They provide estimates on lengths of curves and curvature as one propagates the input along the net's layers. Briefly put, it shown that curvature and lengths could increase exponentially in terms of $L$ thus equipping the random network model with rich expressive capabilities. To provide a more formal perspective of these results, let us define

\begin{equation}
    q^l := \frac{1}{N_l} \sum_{i=1}^{N_l} (\alpha^l_i)^2, \quad q^l_{ab} := \frac{1}{N_l} \sum_{i=1}^{N_l} \alpha^l_i(\s{x}) \alpha^l_i(\s{x}'),
\end{equation}
where $\s{x}, \s{x}'$ are two inputs. Here, $q^l$ captures the vector's length whereas $q_{ab}^l$ describes angles.  
Based on the dynamics of Eq. (\ref{eq:Poole-ff-dynamics}), \cite{poole:neurips:2016} derived expressions on the evolution of length and scalar product (see Proposition \ref{prop:Poole-lengths-covariance} in the Supplementary) in terms of the weights initialization (e.g. $\sigma_w$) and depth $L$. Furthermore, \cite{raghu:icml:2017} refined the analysis and were able to obtain exponential bounds on curve lengths in terms of, e.g., depth $L$, $\sigma_w$ and properties of the non-linearity $\varphi$.

\textbf{Lengths of vectors and scalar products.} Using the result of Theorem $\ref{thm:gp-ff}$ and Proposition ~\ref{prop:srank-covariance} we can obtain the following version of Proposition \ref{prop:Poole-lengths-covariance}, Appendix, in our setting:

\begin{prop}[Supplementary, Section \ref{sec:feed_dynamics}]
\label{thm:srank-lengths-scalar-prods}
    Let us suppose that $\b{W}^l$ are sampled as in Algorithm \ref{alg:naive_sampling} with given target spectral norm $s_l$ and stable rank $r_l$. Then, for sufficiently large layer widths, the feed-forward dynamics of the length $q^l$ and two-input covariance $q^l_{ab}$ is given by
    \begin{align*}
        q^l \approx \frac{r_l s_l^2}{N_l} \Exp_{z} \left[ \varphi(\sqrt{q^{l-1}} z)^2 \right], \quad q^l_{12} \approx \frac{r_l s_l^2}{N_l} \Exp_{z_1, z_2} \left[ \varphi(z_1) \varphi(z_2) \right],
    \end{align*}
    where the distribution $(z_1, z_2)$ is given by the $l-1$-th layer GP dynamics outlined in Theorem \ref{thm:gp-ff}.
\end{prop}

\begin{figure}
    \includegraphics[width=1\linewidth]{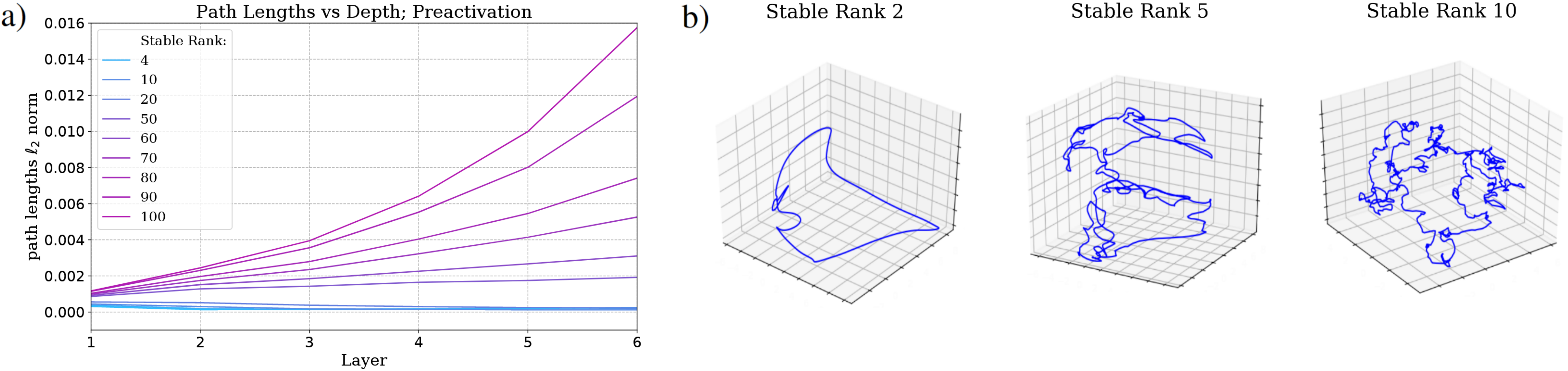}
    \caption{\textbf{a)} Distances between random pairs of MNIST examples in terms of $\ell_2$ norm during propagation through a 6-layer MLP with hidden dimension $h_d$ = 1000. Each curve corresponds to an averaged-out distance between pairs of MNIST samples. As seen in Proposition \ref{thm:srank-lengths-scalar-prods} distances increase with larger stable ranks. \textbf{b)} Propagating a 1d circle through a random 5-layer MLP of hidden dimension 200 initialized via Algorithm \ref{alg:naive_sampling}. Left to right: increasing the layers' stable ranks indicate increasing geometric complexity, lengths and curvature (Propositions \ref{thm:srank-lengths-scalar-prods}, \ref{prop:curvature}). }
    \label{fig:path_distortion}
\end{figure}

In this formulation we explicitly see the influence of stable ranks and spectral norms have to lengths and scalar products. An example of the behaviour of lengths is given in Fig. \ref{fig:path_distortion}.


\textbf{Evolution of curvature.} As outlined in \cite{poole:neurips:2016} the feed-forward dynamics of the length in Proposition \ref{prop:Poole-lengths-covariance} implies that under mild conditions on the non-linearity $\varphi$ there exists a fix-point $q^*$ of Eq. (\ref{eq:Poole-length-dynamics}) in the Supplementary,
\begin{equation}
    q^* \approx \sigma_w^2 \int \varphi(\sqrt{q^{*}}z) Dz,
\end{equation}
that is, the forward pass of the DNN does not alter the size of vectors of length $q^*$ - in other words it is a mapping from the sphere of radius $q^*$ to itself. Now, let us consider an input that is a subset of this invariant sphere of radius $q^*$ - in particular, let us consider a 1d circle of radius $q^*$ on this sphere. We are interested in the evolution of the geometry of the circle into curves as it is propagated down the DNN's layers. At a given point $t$ on a curve, let the osculating circle have a radius $R(t)$. Then we define $\kappa (t) = 1/R(t)$ as the curvature at that point. Furthermore, we denote by $g$ the metric induced on a curve from the ambient Euclidean space. With that in place, we have the following:
\begin{prop}
\label{prop:curvature}
    The feed-forward dynamics of the metric $g^l$ and curvature $\kappa^l$ of the circle are given by 
    \begin{align}
        g^l & = \chi_1 g^{l - 1}, \quad
        \left( \kappa^l\right)^2 = 3\frac{\chi_2}{\chi_1^2} + \frac{1}{\chi_1}\left( \kappa^{l - 1}\right)^2, 
    \end{align}
    where 
    \begin{align*}
        \chi_1   = \frac{(s_t^l)^2 r_t^l}{N_l N_{l-1}}\int \left[\varphi'(\sqrt{q^*}z)\right]^2 Dz, \quad \chi_2  = \frac{(s_t^l)^2 r_t^l}{N_l N_{l-1}}\int \left[\varphi''(\sqrt{q^*}z)\right]^2 Dz.
    \end{align*}
    In particular, the stable rank boost/reduce geometric features exponentially with respect to depth. 
\end{prop}

Thus, as in the case of vector lengths, lower stable ranks yield a decrease in curvature 
(Fig. \ref{fig:path_distortion}).

\section{Training analysis - NTK evolution}
\label{sec:ntk-evolution}

A recent well-known line of work \cite{jacot:neurips:2018, Hanin2020Finite, Yang_2019} studies the training evolution of DNN models in terms of Neural Tangent Kernels. In the following, we show how our non-i.i.d. and non-Gaussian initilization scheme affects the NTK behavior. In short, similarly to the i.i.d. Gaussian case, we obtain a recursive (in terms of layer depth) NTK description where the stable ranks and spectral norm appear explicitly. Furthermore, for large widths we show that the NTKs are deterministic and remain constant during training (as in the Gaussian case). 
We recall that the NTK $\s{\Theta}^L(\s{\theta})$ is defined by:
\begin{equation}\label{def:NTK}
    \s{\Theta}^L(\s{\theta}) = \sum_{p = 1}^P \partial_{\theta_p}\b{F}^L(\s{\theta}) \otimes \partial_{\theta_p}\b{F}^L(\s{\theta}),
\end{equation}
where $\b{F}^L : \mathbb{R}^P \to \mathcal{F}, \s{\theta} \mapsto f_\theta$ represents the evaluation function of the DNN having $L$ layers, where the parameter vector $\s{\theta}$ lies in the Euclidean space $\mathbb{R}^P$, and $\mathcal{F} = \{ f_\theta : \s{\theta} \in \mathbb{R}^P \}$ represents the space of output functions $f_{\s{\theta}}$. For further background we refer to Appendix, Section \ref{subsec:ntk-background}.

\begin{figure}[t]
    \includegraphics[width=1.\linewidth]{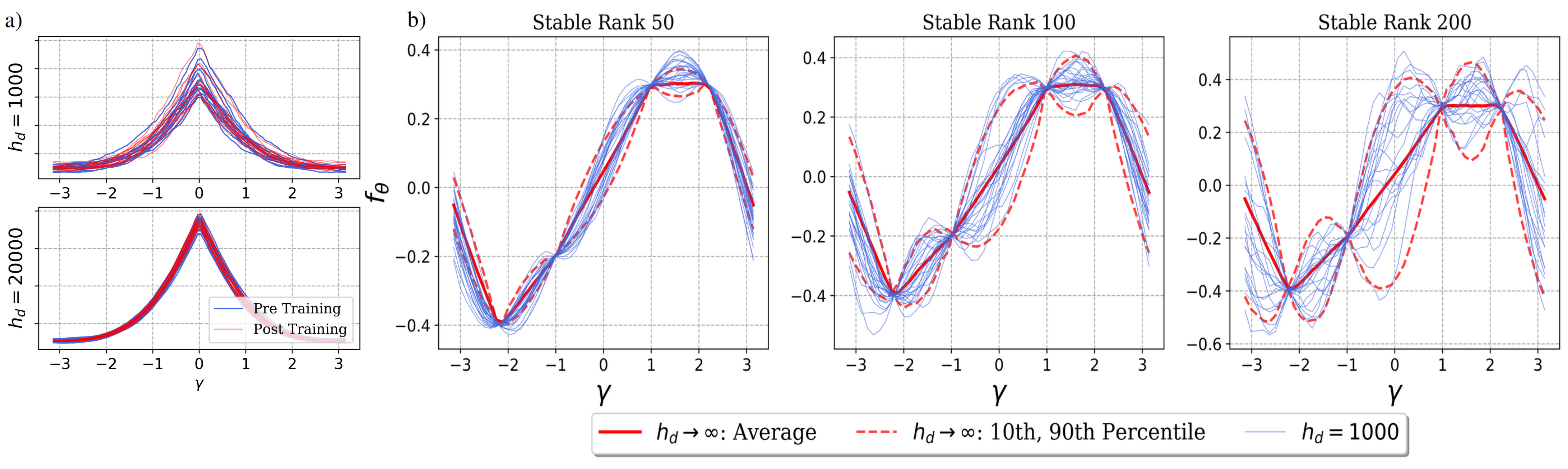}
    \caption{ Convergence properties of NTK for a 5-layer MLP. \textbf{a)} We consider a 1d circle parametrized by angle $\gamma$ and fit $4$ random points. We compare finite-width fitting via least-squares (blue curves) against the large-width GP evolution prescribed by the NTK (red curves) for $10$ initializations. First, the NTK evolution gives a reasonable approximation of the smaller-width least-squares solution. Second, stable ranks clearly influence the variance of the curves as proved in Sections \ref{sec:kernel_methods}, \ref{sec:ntk-evolution}.
    \textbf{b)} We plot NTKs by fixing a first input point and let the second one vary along on a 1d circle parametrized by angle $\gamma$ as before. We compare pre- and post-training NTKs (after 500 steps) for $10$ initializations. Left and right we plot finite vs. large-width behaviour.}
    \label{fig:NTKs}
\end{figure}

\textbf{NTK behavior at initialization.} Now we record the large width behavior of the NTK in our special setting. Our results in this context are parallel to (and inspired from) the main theorems of \cite{jacot:neurips:2018}, which were further adapted to the setting of simultaneous limits in \cite{Yang_2019}. 
In both the sequential and simultaneous limit regimes, we have:
\begin{prop}[NTKs are deterministic, Supplementary, Section \ref{sec:7.1_7.2_7.3}] \label{thm:Thm_1_Jacot}
    In the 
    large width regime, the NTK $\s{\Theta}^L$ converges in probability to a deterministic limiting kernel:
    \begin{equation*}
        \s{\Theta}^L \longrightarrow \s{\Theta}^L_\infty \otimes Id_{N_L}.
    \end{equation*}
    The kernel $\s{\Theta}^L_\infty$ is defined recursively by 
    \begin{align}\label{eq:recursive_NTK}
        \s{\Theta}_\infty^1(\s{x}, \s{x}') & = 
        \left(\gamma^1\right)^2\langle \s{x}, \s{x}'\rangle + 1 \nonumber\\
        \s{\Theta}^{l}_\infty(\s{x}, \s{x}') & = \left(\gamma^l\right)^2\frac{r_ls_l^2}{N_l}\s{\Theta}^{l - 1}_\infty(\s{x}, \s{x}')\Dot{\s{\Sigma}}^{l}(\s{x}, \s{x}') + \s{\Sigma}^{l}(\s{x}, \s{x}'),
    \end{align}
    where $\Dot{\s{\Sigma}}^{l}(\s{x}, \s{x}') = \Exp_{f \sim \mathcal{N}(0, \s{\Sigma}^{l - 1})} \left[ \Dot{\varphi}(f(\s{x}))\Dot{\varphi}(f(\s{x}')) \right].$
\end{prop}

\textbf{NTK behavior during training.} Our sampling scheme assures that at initialization the layer weights possess a certain stable rank and spectral norm. However, this property is not guaranteed to hold along the gradient descent evolution. Indeed, the weights would evolve via the ODE $\partial_t \theta_p(t) = -\langle \partial_{\theta_p} \b{F}^L, \b{d}_t \rangle_{N_0}, $
where $t \mapsto d_t \in \mathcal{F}$ would indicate the training direction. To ensure the invariance of stable ranks along the gradient descent evolution one could use a projection scheme (see also \cite{sanyal:iclr:2020}). More precisely, for $l = 1, \dots, L$, let $\s{\Pi}_l := \s{\Pi}(s_l, r_l)$ denote the ``projection'' of a matrix onto the tangent space of the class of matrices of stable rank $r_l$ and spectral norm $s_l$. Then, one can consider the constrained gradient descent given by the differential equation:
\begin{equation*}
    \partial_t \s{\theta} (t) = \s{\Pi}\left( \langle \partial_{\theta_1} \b{F}^L, \b{d}_t\rangle_{N_0}, \dots, \langle\partial_{\theta_P} \b{F}^L, \b{d}_t \rangle_{N_0}\rangle \right),
\end{equation*}
where $\s{\Pi}$ is the operator that acts on the 
$l$-th layer by $\s{\Pi}_l$. 
We now state the asymptotic behavior of the NTK during both plain or constrained (via stable rank projection) training for the {\em sequential limit}:
\begin{prop}[NTKs are constant during training, Supplementary, Section \ref{sec:7.1_7.2_7.3}]\label{thm:large_width_training}
    Suppose that the integral $\int_0^T \| \b{P} \b{d}_t \|_{N_0}\; dt$ 
    is stochastically bounded, where $\b{P}$ is either the identity operator or the projection operator $\s{\Pi}$ as above. For any such $T$, in the limit $N_1, \dots, N_{L - 2} \to \infty$ sequentially,  we have, uniformly for $t \in [0, T]$, 
    \begin{equation*}
        \s{\Theta}^L(t) \longrightarrow \s{\Theta}^L_\infty \otimes Id_{N_L}.
    \end{equation*}
    In the limit, the dynamics of $f_{\s{\theta}}$ is described by the differential equation $\partial_t f_{\s{\theta}(t)} = \Phi_{\s{\Theta}^L_\infty \otimes Id_{N_L}}\left( \langle \b{P} \b{d}_t, .\rangle_{N_0}\right).$
\end{prop}
Numerics on NTK convergence pre/post-training and kernel regression are presented in Fig. \ref{fig:NTKs}.

\textbf{Eigenspectrum of the constrained stable rank NTK.} 
The  eigenspectrum can play a significant role in determining
whether the network fits clean label faster than noisy labels as optimization will be faster (slower) for larger (lower respectively) eigenvalues \cite{Arora_fine_grain_ICML, Oymaketal}. 
It is natural that stable rank normalization lowers the rank of the kernel, thereby preventing the fitting of noise as observed both in Appendix, Section \ref{sec:further-emp-results} and in \cite{sanyal:iclr:2020}. We have the following (see Supplementary Section  \ref{subsec:eigenspectrum} for a proof): 
\begin{prop}\label{prop:eigenspectrum_NTK}
For constrained stable rank DNNs where $s_l^2r_l \leq  C\frac{N_l}{(\gamma^l)^2}$, the eigenspectrum of the NTK becomes arbitrarily small in the large width regime as $C$ becomes arbitrarily small.
\end{prop}

\section{Further empirical results}
\label{sec:experiments}

\begin{figure}
    \centering
    \begin{center}
        \includegraphics[width=0.7\linewidth]{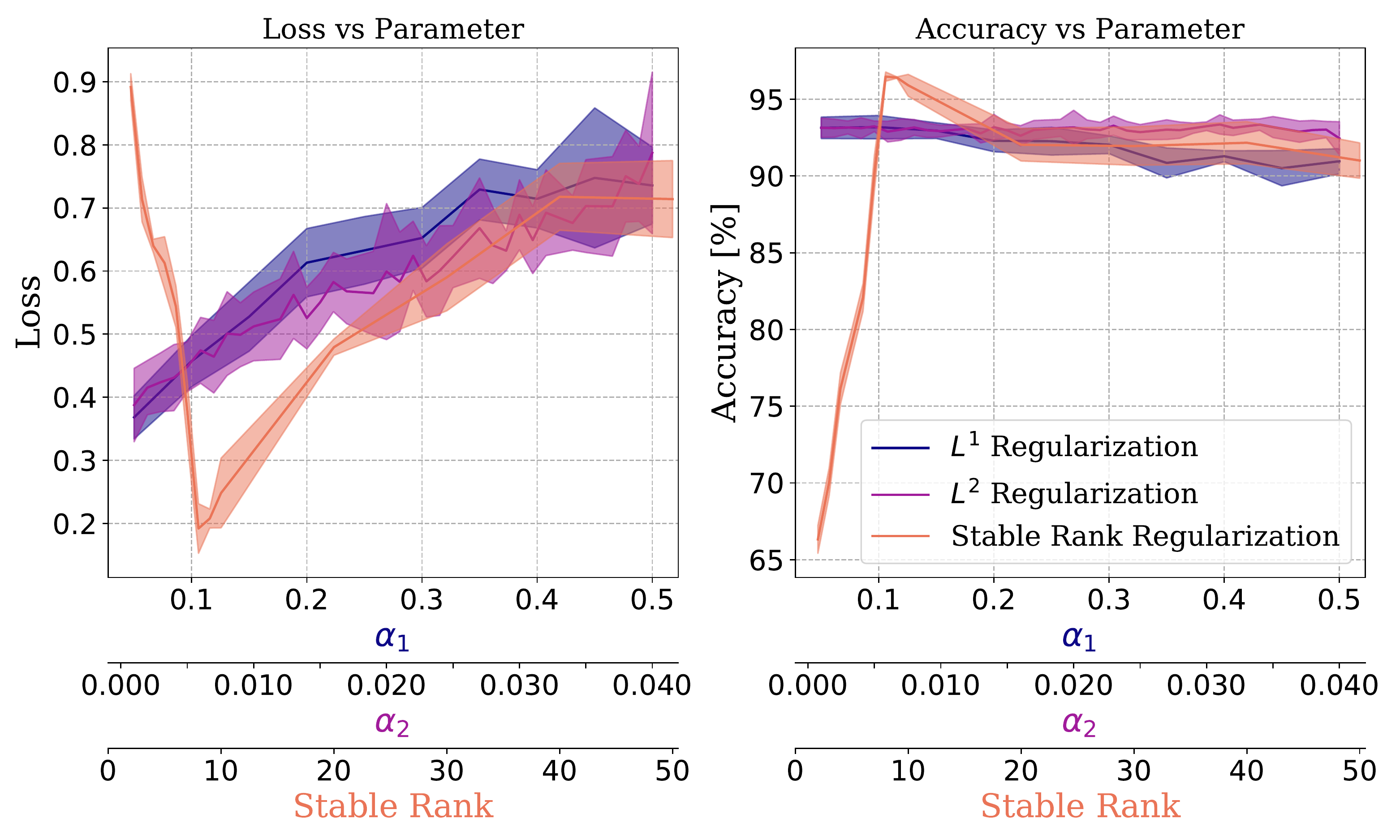}
    \end{center}
    \caption{Comparison of regularization via $L^1$, $L^2$ and our stable rank initialization scheme: loss and accuracy versus the respective hyper-parameter. For each hyper-parameter value we trained (via SGD) 100 3-layer-MLP models (with 750 nodes per hidden layer) on the MNIST dataset for one full training epoch; afterwards, using the test set, we evaluated the test loss/accuracy shown above.}
    \label{fig:reg-comparison}
\end{figure}


\textbf{Training speeds and generalization.} A related question arises as to stable rank initialization and stable rank projection (as in Subsection NTK) necessarily leads to improvements in convergence speeds and model generalization. First, stable rank projection (i.e. normalize stable rank after each training step) has been thoroughly investigated in \cite{sanyal:iclr:2020} and improved generalization properties have been reported. Although not explicitly reported, it is natural to expect that such a projection would impede training speeds - to this end, in the Appendix Figure \ref{fig:conv-speed} for completeness we provide an illustrative explicit example as to how geometrically convergence could be slowed down. Intuitively, we consider a simple least-squares convex problem where stable rank control forces the gradient descent curves to be projected on certain spherical hypersurfaces whereas the plain gradient-descent trajectories follow optimal straight lines. Numerical outline is given in Fig. \ref{fig:conv-speed} with further details provided in Supplementary, Section \ref{subsec:training-speeds}.

Another intriguing problem is whether initialization alone assists training speed and generalization. Basic experiments and intuition suggest that a given dataset poses conditions on what stable rank the class of optimal solutions are allowed to attain - e.g., this is naively manifested in linear regression problems, where the data explicitly determines the weights (and hence, their stable ranks); for further discussion we refer to Appendix, Section \ref{sec:further-emp-results}. In a more general DNN situation one might similarly expect that appropriate stable rank initialization would focus the solution space and would lead to improved convergence and generalization. After experimenting with synthetic data (Appendix, Section \ref{sec:further-emp-results}) we collected some results on the MNIST dataset in Fig. \ref{fig:reg-comparison} where we directly compare the proposed stable rank initialization with other regularization schemes such as $L^1, L^2$ that have proven effective in decreasing the model's complexity and geometry \cite{Serra2020}. Roughly, in Fig. \ref{fig:reg-comparison} we observe that depending on the relevant hyperparameter setup each regularization scheme has a certain "sweet-spot" regime - what is more, the initialization with stable rank $\sim 10$ provides a certain advantage in terms of convergence and generalization. In other words, although simplistic, these experiments suggest that initialization with the "appropriate" stable rank is beneficial - an exciting remaining problem is how one could a priori estimate the stable rank constraints of this optimal regime in terms of the given data, training algorithm and model architecture. Further analysis of the relation between noise fitting and stable rank constraints is provided in Appendix Figure \ref{fig:shuffled_mnist}.

\section{Conclusions}
\label{sec:conclusion}

We have analyzed the effect of layer weights with controlled stable rank and spectral norm in deep neural networks. After proposing a suitable sampling scheme, we studied properties of the DNN at initialization as well as during training. This led to results on geometric expressivity (lengths, scalar products, curvature), generalization trade-off, NTK training description and empirical evaluation. In particular, theoretically we have made use of non-trivial central limit theorems associated with the group of orthogonal matrices $O(n)$ as well as Weingarten moment estimates; further, we have provided an empirical evaluation of convergence illuminating how stable rank initialization/regularization could prove beneficial. As a consequence, our analysis demonstrates that stable ranks can be used to obtain smoother and less overfitting models with theoretically improved generalization errors. We hope the present study aids and further motivates the understanding of compression-based or parameter reduction initialization/regularization schemes.

\section{Dynamics with Gaussian initialization}
\label{sec:gaussian-proofs}
We begin by recalling the following result regarding the feed-forward dynamics of Gaussian i.i.d. neural nets:

\begin{prop}[\cite{poole:neurips:2016}] \label{prop:Poole-lengths-covariance}
    Let the weights $W^l_{ij}$ be drawn i.i.d. from a zero mean Gaussian with variance $\sigma_w^2/N_{l-1}$, while the biases are drawn i.i.d. from a zero mean Gaussian with variance $\sigma_b^2$. Provided the network widths $N_l$ are sufficiently large, the following recursive relations for $q^l$ hold:
    \begin{equation} \label{eq:Poole-length-dynamics}
        q^l \approx \sigma_w^2 \int \varphi(\sqrt{q^{l-1}}z) Dz,
    \end{equation}
    where $Dz$ denotes the standard Gaussian measure $\frac{e^{-z^2/2}}{\sqrt{2\pi}} dz$ and $N_{l}$ is assumed suffiently large. Further, for the covariance $q^l_{ab}$ one has:
    \begin{equation}
        q_{12}^l \approx \sigma^2_w \int \varphi(u_1) \varphi(u_2) Dz_1 Dz_2 + \sigma_b^2, \text{ where}
    \end{equation}
    \begin{equation}
        u_1 = \sqrt{q_{11}^{l-1}}, ~ u_2 = \sqrt{q_{22}^{l-1}} \left( c_{12}^{l-1} z_1 + \sqrt{1 - (c_{12}^{l-1})^2} z_2 \right).
    \end{equation}
\end{prop}

\begin{proof}[Sketch of Proof]
    Using the definitions and denoting the empirical expectation by $\Expemp$, one writes
    \begin{align*}
        q^l &= \Expemp \left[  (\alpha^l_i)^2 \right] = \Expemp \left[ (\s{W}^{l}_{i, \cdot} \cdot \varphi(\s{\alpha}^{l-1}))^2 \right] + \sigma_b^2 \\ &= \frac{\sigma_w^2}{N_{l-1}} \sum_{i=1}^{N_{l-1}} \varphi(\alpha_i^{l-1})^2 + \sigma_b^2  \approx \sigma_w^2 \int \varphi(\sqrt{q^{l-1}}z)^2 Dz + \sigma_b^2 ,
    \end{align*}
0    where $Dz$ denotes the standard Gaussian measure $\frac{e^{-z^2/2}}{\sqrt{2\pi}} dz$ and $N_{l-1}$ is sufficiently large so that the integral approximation is sufficiently close by the central limit theorem. Here one uses essentially that
    \begin{equation} \label{eq:W-covariance}
        \Expemp \left[ \s{W}^{l}_{j, \cdot }, \s{W}^{l}_{k, \cdot} \right] \approx \frac{\delta_{jk} \sigma^2_w}{N_{l-1}}.
    \end{equation}
    
    The computation for $q^l_{ab}$ is similar and we refer to \cite{poole:neurips:2016} for further details.

\end{proof}

\section{Sampling properties}\label{sec:Sampling_properties}

Proof of Proposition \ref{prop:srank-mean-field-nn} of the main text.
\begin{proof}
    First we recall that for random matrices whose entries are zero mean i.i.d. variables whose variance is $1$ and the fourth moment is bounded \cite{bai1993, Rudelson2010}, the following convergence almost surely holds:
    \begin{equation}
        \frac{1}{\sqrt{N_l}} \sigma_1 \rightarrow 1 + \sqrt{\alpha}.
    \end{equation}
    In our case, the matrix $\sqrt{N_{l-1}} \s{W}^l / \sigma_w$ satisfies the above conditions so for the largest singular eigenvalue $\sigma_1^{\s{W}}$ of $\s{W}^l$ we obtain the almost sure convergence:
    \begin{equation}
        \frac{1}{\sigma_w}\sqrt{\frac{N_{l-1}}{N_l}} \sigma_1^{\s{W}} = \frac{\sqrt{\alpha}}{\sigma_w} \sigma_1^{\s{W}} \rightarrow 1 + \sqrt{\alpha}.
    \end{equation}
    
    Furthermore, by construction
    \begin{align}
        \|\s{W}^l\|_F^2 &= \sum_{i, j} \left(W^l_{ij}\right)^2 = \sum_{i=1}^{N_{l-1}} N_{l} \left( \frac{1}{N_{l}} \sum_{j=1}^{N_{l}} \left(W^l_{i,j}\right)^2 \right) \\ 
        &\approx \sum_{i=1}^{N_{l-1}} N_{l} \frac{\sigma_w^2}{N_{l-1}} = N_{l} {\sigma_w^2} 
    \end{align}
    where we have used that $N_l$ is sufficiently large and the empirical second moment approximates almost surely the true one. Putting the above convergence estimates together finishes the claim.
\end{proof}

\subsection{Correctness and success rates}
\label{subsec:correctness}
Proof of Proposition \ref{prop:samp_prod} of the main text.

\begin{proof}
    First, it follows from the construction that the maximal eigenvalue $s_1$ of the output is $s_t$. 
    Furthermore, using the definition of stable rank, it follows that $\srank(S) = r_t$.
    Geometrically, the space of vectors $\{s_i\}$ that induce the required matrices is the intersection $[0, s_t]^{m-1} \cap (s_t \sqrt{r_t - 1}) \mathbb{S}^{m-2}$, where $\sqrt{r_t - 1}$ is the scaling factor from the algorithm. It is well-known that normalizing standard Gaussian vectors produces a uniform distribution on the sphere, hence, after Step $4$, 
    $\{s_i\}_{i=2}^{m}$ are uniformly distributed over the sphere $\mathbb{S}^{m-2}$ in the positive quadrant. Finally, Step 
    $5$ just takes the intersection with the $m-1$-dimensional cube and, hence, produces the required uniform distribution over $[0, s_t]^{m-1} \cap (s_t \sqrt{r_t - 1}) \mathbb{S}^{m-2}$.
\end{proof}
Now we give a proof of Proposition \ref{prop:gaussian-acceptance-rate} of the main text.
\begin{proof}
    In order to accept a sample we need Step $5$ in Algorithm~\ref{alg:naive_sampling} of the main text 
    to be true. 
    In practice, we need $\frac{\sqrt{r_t - 1}}{\sqrt{\sum_{i=2}^m s_i^2}}|s_i| \leq 1,~i=2,\dots,m$, so we can decompose this bound as $\frac{\sqrt{r_t - 1}}{\sqrt{\sum_{i=2}^m s_i^2}} \leq \eta$ and $\eta \leq \frac{1}{|s_i|}$. Clearly, when both conditions hold, a sample is accepted. Therefore, we can compute the corresponding probability as
    \begin{align}
        \Prob\left[ \text{Accept} \{ s_i \} \right] \geq \Prob \left[ \left( \sum_{i=2}^m s^2_i \geq \frac{r_t - 1}{\eta^2} \right) \bigcap \left( \bigcap_{i=2}^m |s_i| \leq \frac{1}{\eta} \right)  \right]
    \end{align}
    where $\eta$ is an arbitrary real positive number. Hence, by using that $\Prob[A \cap B] = 1 - \Prob[\bar{A} \cup \bar{B}] \geq 1 - \Prob[\bar{A}] - \Prob[\bar{B}] $, one gets
    \begin{equation} \label{eq:lower-bound-acceptance}
        \Prob\left[ \text{Accept } \{ s_i \} \right] \geq 1 - \Prob\left[ \sum_{i=2}^m s_i^2 < \frac{r_t - 1}{\eta^2} \right] - \sum_{i=2}^m \Prob \left[ |s_i| > \frac{1}{\eta} \right].
    \end{equation}
    The last term on the right is clearly equal to $(m-1)\left( 1 - 2 \erf(\frac{1}{\eta})\right)$. We observe that $\sum_{i=2}^m s_i^2$ follows a chi-squared law with $m-1$ degrees of freedom. The second term on the right hand side in Eq.~\ref{eq:lower-bound-acceptance} is precisely the corresponding cdf $F_{m-1}(x)$ evaluated at $x = (r_t - 1)/ \eta^2$. Appropriate upper bounds on the tails \cite{Sanjoy2003} yield for $x \leq m-1$
    \begin{equation}
        F_{m-1} (x) \leq \left(z e^{1 - z} \right)^{\frac{m-1}{2}}, \quad z = \frac{x}{m-1} \in [0, 1].
    \end{equation}
    Hence, to apply the bound one needs $\eta^2 \geq (r_t - 1) / (m - 1)$. Putting the above two estimates together yields the claim.
\end{proof}

\subsection{Further sampling methods}
\label{subsec:further-sampling}

While the procedure proposed in Subsection \ref{subsec:uniform-lebesgue-sampling} of the main text starts with vectors uniformly sampled on a sphere and then intersects with a cube, another approach would be to start with samples uniformly distributed in a cube and then normalize to a sphere. This method puts more emphasis on spectral space uniformization; however, due to the uniform sampling and normalization afterwards, the produced matrix samples will in general not be uniformly distributed w.r.t. surface measure as in the first scheme in Subsection \ref{subsec:uniform-lebesgue-sampling} of the main text. 
We outline the method in Algorithm~\ref{alg:alternate}.

\begin{algorithm}[!ht]
	\caption{Sampling uniformly in a cube}\label{alg:alternate} 
	\label{alg:improved_sampling}
        \begin{algorithmic}[1]
         	\REQUIRE{Input}
         	\ENSURE{Output}
         	\STATE{Set $s_1 := 1$, $m := \min(N_l, N_{l-1})$}
         	\STATE{Set $s_1 := 1$, $m := \min(N_l, N_{l-1})$}
         	\STATE{Set $s_i \sim \mathcal{U}([0, 1]) $ for $i = 2, \dots, m$}
         	\STATE{Define $r := \frac{\sqrt{r_t - 1}}{ \sqrt{\sum_{i=2}^{m} s_i^2}} $}
         	\STATE{If $r \leq 1$, accept the sample, rescale $s_i := r * s_i, i = 2, \dots, m$ and return $S := diag(s_t * \{s_i\}_{i=1}^m) \in \mathbb{R}^{N_l \times N_{l-1}}$. Else, go to Step 2.}
        \end{algorithmic}
\end{algorithm}


\begin{prop}
    The above algorithm produces a matrix with stable rank $r_t$ and spectral norm $s_t$.
\end{prop}

\begin{proof}
    One directly checks using the definitions of stable rank that the output matrix is having the required properties. Similarly to Proposition \ref{prop:gaussian-acceptance-rate} of the main text one can obtain lower bounds on the acceptance rate of the algorithm.
\end{proof}

\section{
Large width asymptotics of the random stable rank DNN}\label{sec:proof_large_width_single_layer}\label{sec:feed_dynamics}
Before going into the proof of Lemma \ref{lem:large_width_behavior} of the main text, we first prove the following technical preliminary:
\begin{proof}[Proof of Proposition \ref{prop:srank-covariance} of the main text]
    Using the singular value decomposition and independence one obtains
    \begin{align*}
        \Exp \left[ W^l_{i, j} W^l_{p, q} \right] &= \Exp \left[ \sum_{a, b} U_{ia} U_{pb} \Sigma_{aa} \Sigma_{bb} V_{aj} V_{bq} \right] \\  &= \sum_{a, b} \Exp \left[ U_{ia} U_{pb} \right] \Exp \left[ \Sigma_{aa} \Sigma_{bb} \right] \Exp \left[ V_{aj} V_{bq} \right].
    \end{align*}

    To handle the first and third terms one could use well-known integration results for polynomials over the orthogonal group equipped with the Haar measure \cite{Collins2006} in terms of Weingarten functions. One obtains
    \begin{equation}
        \Exp\left[ U_{ia} U_{pb} \right] =  \delta_{ip} \delta_{ab} \frac{1}{N_l}, \quad \Exp\left[ V_{aj} V_{bq} \right] =  \delta_{ab} \delta_{jq} \frac{1}{N_{l-1}}.
    \end{equation}
    
    Hence,
    \begin{equation}
         \Exp \left[ W^l_{i, j} W^l_{p, q} \right] = \begin{cases} 0, \quad \text{for } i \neq p \text{ or } j \neq q, \\ \frac{1}{N_l N_{l-1}} \sum_{n=1}^{m} \Exp \left[ s_{nn}^2 \right] = \frac{s_t^2 r_t}{ N_l N_{l-1}}, \quad \text{else},
        \end{cases}
    \end{equation}
    where, finally, we have used that $\boldsymbol{\Sigma}$ possesses a stable rank $r_t$. The second equation follows by noting that, by construction, $s_{11} = s_t$ and $ \sum_{n=1}^m s^2_{nn}/s^2_{11} = r_t$ are fixed, so taking the expectation yields the result.
\end{proof}
Now we start with the proof of Proposition \ref{thm:srank-lengths-scalar-prods} of the main text. As part of the proof, we will also include a proof of the important technical result Proposition \ref{lem:large_width_behavior} of the main text. 
\begin{proof}
First, we start with the initial layer as a base case and consider the pre-activations
    
    \begin{equation}
        f_{\s{\theta}} (\s{x}) = \s{W}^1 \s{x} + 
        \s{b}^1, \quad f_{\s{\theta}, k}(\s{x}) = \sum_{i=1}^{N_0} W^1_{ki} x_i + 
        b^1_k.
    \end{equation}
    Let us consider for simplicity the bias-free case and set $
    \s{b}^1$ to $\s{0}$. 
    
    In general, the marginal distribution of the column vector $\s{W}_1$ is not the same as $\s{W}_1 | \s{W}_2, \dots \s{W}_{N_1}$ since $\|\s{W}_1\|^2$ is deterministically given by the stable rank of $\s{W}$, the spectral norm (which gives the scaling) and the norms of all the other rows.
    
    First, we prove a couple of technical lemmas.
    \begin{lem} The output functions $f_{\s{\theta}, k}(\s{x})$ are centered, identically distributed and dependent. Furthermore, $\Exp \left[ \|f_{\s{\theta}}(\s{x})\|^2 \right] = r_0 s_0^2 \|\s{x}\|^2 / N_0$ and the covariance is given by $\Exp\left[ f_{\s{\theta}, k} (\s{x}) f_{\s{\theta}, k} (\s{x}') \right] = r_0 s_0^2 \langle \s{x}, \s{x}' \rangle / N_0$.
    \end{lem} 
    
    \begin{proof}[Proof of Lemma] 
    First, one can express $f_{\s{\theta}, k}(\s{x}) = \s{e}_k \s{W} \s{x} = \s{W}_k \s{x}$ and observe that by construction $\s{e}_k \s{W}$ follows the same distribution as $\s{e}_l \s{W}$ for each pair of indices $k, l$ since $\s{e}_k U^T$ and $\s{e}_l U^T$ are uniformly distributed on the unit sphere.
    

    Finally, using the properties of $\s{W}$ we have
    \begin{align*}
        \Exp \left[ \| f_{\s{\theta}} (\s{x}) \|^2 \right] & = \Exp \sum_{k=1}^{N_1} f_{\s{\theta}, k}(\s{x})^2 = \Exp \sum_{i, k} \s{W}_{ki}^2 \s{x}_i^2 \\ & = N_1 \frac{s_0 r_0^2}{N_1 N_0} \|\s{x}\|^2.
    \end{align*}
    This concludes the proof of the Lemma.
\end{proof}
Next, we study the distribution of $(z_1, z_2) := \left(f_{\s{\theta}, k}(\s{x}), f_{\s{\theta}, k}(\s{x}') \right) $ for two given inputs $\s{x}, \s{x}' \in \mathbb{R}^{N_0}$.
    
    \begin{lem}
        The joint distribution of $(z_1, z_2) \sim \omega$ is completely determined by $\|\s{x}\|, \|\s{x}'\| $ and the scalar product $\langle \s{x}, \s{x}' \rangle$.
    \end{lem}
    
    \begin{proof}[Proof of Lemma]
        By definition we have $(z_1, z_2) = \left(\sum_i W_{ki} x_i, \sum_j W_{kj} x_j'\right)$. Hence the pointwise density of $(z_1, z_2)$ can be computed by accumulating the density of $\s{W}$ over the instances where the following system holds:
        \begin{equation}
            \begin{cases}
                \sum_iW_{ki} x_i = z_1, \\
                \sum_j W_{kj} x_j' = z_2.
            \end{cases}
        \end{equation}
        Since $\s{W}$ and $\s{W}\s{R}$ for a rotation matrix $\s{R}$ follow the same distribution, we can arrange that $\s{x} = \|\s{x}\| \s{e}_1$ and $\s{x}'$ is spanned by $\s{e}_1, \s{e}_2$. One straightforwardly checks that the linear system translates to
        \begin{equation}
            \begin{cases}
                W_{k1} = \frac{z_1}{\|\s{x}\|}, \\
                W_{k2} = \frac{\|\s{x}\|}{\sqrt{\|\s{x}'\|^2 - \langle \s{x}, \s{x}' \rangle^2}} \left( z_2 - \frac{\langle \s{x}, \s{x}' \rangle}{ \|\s{x}\|^2} z_1 \right).
            \end{cases}
        \end{equation}
        Thus we obtain the density of $(z_1, z_2)$ as $p_{s, r}(\s{W} | W_{k1}, W_{k2})$ where the conditioning depends only on 
        $\|\s{x}\|, \|\s{x}'\|, \langle \s{x}, \s{x}' \rangle$. This completes the proof of the Lemma.
    \end{proof}

The next result is at the technical heart of much of our paper.
\begin{proof}[Proof of Proposition \ref{lem:large_width_behavior} of the main text]
    The main tool we employ is a Central Limit Theorem (CLT) result for Haar distributed orthogonal matrices:
    \begin{thm} [\cite{aristotile03, Meckes08, Fulman2012}] \label{thm:aristotle}
        For some $n \in \mathbb{N}$, let $\s{A}_n \in \mathbb{R}^{n \times n}$ be a deterministic matrix such that $\tr(\s{A}_n \s{A}_n^T) = n$ and let $\s{U}_n \in O(n)$ be uniformly distributed according to the Haar measure on the $n$-dimensional orthogonal group $O(n)$. Then $\tr(\s{A}_n \s{U}_n)$ converges in distribution to $N(0, 1)$ as $n \rightarrow \infty$. Moreover, the convergence rate is of order $1 / (n-1)$ in total variation distance.
    \end{thm}
    
    First, we address the regime when $N_0$ is large. In this case we can write
    \begin{align*}
        f_{\s{\theta}, k}(\s{x}) & = \sum_j W_{kj} x_j + b_k = \sum_{a, j} U_{ka} 
        S_{aa}
        V_{ja} x_j + b_k \\
        & = \sum_{a, j} V_{aj} h_a x_j + b_k, 
    \end{align*}
    where $\s{U} \in O(N_1), \s{V} \in O(N_0)$ and where we have denoted $h_a = U_{ka} S_{aa}$. Further, we can write the latter as
    \begin{equation}
        f_{\s{\theta}, k}(\s{x}) = \tr(\s{V} (\s{h} \otimes \s{x})) + b_k.
    \end{equation}
    Let us now fix $\s{U}, \s{S}$ and $\s{x}$. From Theorem \ref{thm:aristotle} and the independence between the bias and the layer weights it follows that
    \begin{equation}
        \frac{\sqrt{N_0}}{\| \s{h} \otimes \s{x} \|} f_{\s{\theta}, k} (\s{x}) \rightarrow_d N \left(0, 1 + \frac{N_0 \sigma_b^2}{\|\s{h} \otimes \s{x} \|^2} \right), \quad N_0 \rightarrow \infty.
    \end{equation}
    The right-hand side has a non-exploding variance provided we select 
    \begin{equation}\label{eq:sigma_choice}
        \sigma_b^2 \in O(\|\s{h} \otimes \s{x} \|^2 / N_0).
    \end{equation}
    Further, we have
    \begin{equation}
        \|\s{h} \otimes \s{x} \|^2 = \sum_{a, j} \left( U_{ka} S_{aa} x_j \right)^2 = \|\s{x}\|^2 \sum_a \left(U_{ka} S_{aa} \right)^2.
    \end{equation}
    Now, we claim that $N_1 \sum_a \left(U_{ka} S_{aa} \right)^2 $ converges in probability to $\|\s{S}\|^2$ as $N_1$ gets increasingly larger. We note that
    \begin{equation}\label{exp_sum_S}
        \Exp\left[ N_1 \sum_a \left(U_{ka} S_{aa} \right)^2 \right] = \|\s{S}\|^2,
    \end{equation}
    so, in particular, since we have $\|\s{S}\|^2$ fixed independent of the sampling of $\s{S}$,
    \begin{equation}
        \Exp\left[ N_1 \sum_a \left(U_{ka} S_{aa} \right)^2 \right] = \Exp\left[ N_1 \sum_a \left(U_{ka} S_{aa} \right)^2 | S \right] = \|\s{S}\|^2. 
    \end{equation}
    In other words, to prove the above convergence one needs a form of the weak law of large numbers. Using Chebyshev's bound and conditioning on $S$, we have
    \begin{align*}
        \Prob\left[ \left|N_1 \sum_a \left(U_{ka} S_{aa} \right)^2 - \|\s{S}\|^2 \right| > \epsilon \big| \s{S} \right] & 
        \leq N_1^2 \vari\left[ \sum_a \left(U_{ka} S_{aa} \right)^2 \big| \s{S} \right] \\
        &\leq N_1^2 \sum_{a, b} \cov\left[ \left(U_{ka} S_{aa} \right)^2, \left(U_{kb} S_{bb} \right)^2 \big| \s{ S} \right] \\
        &= N_1^2 \sum_{a, b} S_{aa}^2 S_{bb}^2 \left[ \Exp\left[ U_{ka}^2 U_{kb}^2 \right] - \Exp\left[ U_{ka}^2\right] \Exp \left[ U_{kb}^2 \right] \right] \\
        &= N_1^2 \sum_{a, b} S_{aa}^2 S_{bb}^2 \left[ \Exp\left[ U_{ka}^2 U_{kb}^2 \right] - \frac{1}{N_1^2} \right],
    \end{align*}
    where in the last equality we have used the orthogonal Weingarten function for $U_{ka}^2$, \cite{Collins2006}. Further, using the Weingarten functions there are two cases for the first term:
    \begin{equation}
        \Exp\left[ U_{ka}^2 U_{kb}^2 \right] = \begin{cases}
            \frac{3}{N_1(N_1 +2)}, \quad a = b, \\
            \frac{1}{N_1(N_1 +2)}, \quad a \neq b.
        \end{cases}
    \end{equation}
    Hence, for some universal constant $C > 0$ we have the bound:
    \begin{align*}
         N_1^2 \sum_{a, b} S_{aa}^2 S_{bb}^2 \left[ \Exp\left[ U_{ka}^2 U_{kb}^2 \right] - \frac{1}{N_1^2} \right] 
        & \leq C \left( \sum_a S_{aa}^4 + \frac{1}{N_1} \sum_{a \neq b} S_{aa}^2 S_{bb}^2 \right) \\
        &\leq C \left( \|\s{S}\|_4^4 + \frac{\|\s{S}\|_2^4}{N_1} \right).
    \end{align*}
    The last expression depends only on the $L^4$-norm of $S$ since the $L^2$-norm is held fixed by construction. To complete the convergence claim we need to show that in our setup, with large probability, the $L^4$-norm of $S$ decays with larger values of $N_1$. Indeed, for large dimension our sampling of $S$ is close to sampling $S_{aa}$ as independent standard Gaussians and normalizing them by $r_t / \sqrt{N_1}$, so that the target stable rank is achieved. In particular, one can estimate the fourth moments of $S_{aa}$ as $\approx 3 r_t^4 / N_1^2$ which implies that with large probability (increasing with $N_1$)
    \begin{equation}
        \|\s{S}\|_4^4 = \sum_a S_{aa}^4 \leq \frac{3 C' r^4}{N_1},
    \end{equation}
    for some universal positive constant $C'$. Let us denote this event by $A$. We have
    \begin{align*}
        \Prob\left[ \left|N_1 \sum_a \left(U_{ka} S_{aa} \right)^2 - \|\s{S}\|^2 \right| > \epsilon \right] & = \Prob\left[ \left( \left|N_1 \sum_a \left(U_{ka} S_{aa} \right)^2 - \|\s{S}\|^2 \right| > \epsilon \right) \cap A \right]\notag \\ &+ \Prob\left[ \left( \left|N_1 \sum_a \left(U_{ka} S_{aa} \right)^2 - \|\s{S}\|^2 \right| > \epsilon \right) \setminus A \right] \\ &\leq \Prob\left[ \left( \left|N_1 \sum_a \left(U_{ka} S_{aa} \right)^2 - \|\s{S}\|^2 \right| > \epsilon \right) | A \right] \Prob\left[ A \right]  + \Prob\left[ \bar{A} \right]\\& \leq \frac{C^{''}}{N_1},
    \end{align*}
    for some universal positive constant $C^{''}$. This completes the proof of the convergence claim, namely, $N_1 \sum_a \left(U_{ka} S_{aa} \right)^2 $ converges in probability to $\|\s{S}\|^2$ as $N_1$ gets increasingly larger.
    
    Now, we address the regime when $N_1$ is large. Analogously, we can write 
    \begin{equation}
        f_{\s{\theta}, k}(\s{x}) = \tr\left( \s{U} \s{L}\right) + b_k,
    \end{equation}
    where $\s{L}$ is the matrix with all columns $0$ except $L_{ak} = \sum_{b, j} S_{ab}V_{bj}x_j, a = 1, \dots, N_1$. Conditioning on $\s{S}, \s{V}, \s{x}$ and applying Theorem \ref{thm:aristotle} yields
    \begin{equation}
        \frac{\sqrt{N_1}}{\|\s{L}\|} f_{k, \s{\theta}} (\s{x}) \rightarrow_d N\left(0, 1 + \frac{N_1 \sigma_b^2}{\|\s{L}\|^2}\right), \quad N_1 \rightarrow \infty.
    \end{equation}
    To assure a non-exploding variance on the RHS we choose $\sigma_b^2 \in O(\|\s{L}\|^2 / N_1)$. Further, as in the previous case that the following convergence in probability holds:
    \begin{equation}
        N_0 \|\s{L}\|^2 \rightarrow \|\s{x}\|^2 \|\s{S}\|^2, \quad N_0 \rightarrow \infty.
    \end{equation}
    This shows that taking the limits in $N_0$ after $N_1$ and in $N_1$ after $N_0$ yield the same Gaussian distribution for $f_{\s{\theta}, k}(\s{x})$.
    
    Finally, we observe that in the large width limit $f_{\s{\theta}, k}(\s{x})$ is not only Gaussian, but behaves as a Gaussian process in $\s{x}$. More precisely, let $\{\s{x}_s\}_{s=1}^{s_0}$ be an arbitrary finite collection of input vectors. By definition, the $f_{\s{\theta}, k}(\cdot)$ is a Gaussian process provided the distribution of the vector $(f_{\s{\theta}, k}(\s{x}_1), \dots, f_{\s{\theta}, k}(\s{x}_{s_0})) $ is a multivariate Gaussian. This is equivalent to requiring that every linear combination of the coordinates is Gaussian:
    \begin{equation}
       \sum_{s=1}^{s_0} \alpha_s f_{\s{\theta}, k}(\s{x}_s) = \s{W} \sum_{s=1}^{s_0} \left(\alpha_s \s{x}_s\right) + \sum_{s=1}^{s_0} \alpha_s b_k.
    \end{equation}
    By linearity, this has been reduced to the case of a single input vector for which we already know the in the large-width limit the behaviour is Gaussian.
    Now, it remains to compute the covariance function $\s{\Sigma}(\s{x}, \s{x}')$. Note that $f_{\s{\theta}, k}(\s{x})$ is universally bounded by the spectral norm of $\s{W}$ and $\|\s{x}\|$ and hence by the dominated convergence theorem for convergence in distribution (and the continuous mapping theorem for $g(f_{ \s{\theta}, k}(\s{x}), f_{\s{\theta}, k}(\s{x}')) := f_{\s{\theta}, k}(\s{x}) f_{\s{ \theta}, k}(\s{x}')$ we have
    \begin{align*}
         \lim_{N_0, N_1 \rightarrow \infty} \Exp\left[ f_{\s{ \theta}, k} (\s{x}) f_{\s{ \theta}, k} (\s{x}') \right]        & = \Exp \left[ \lim_{N_0, N_1 \rightarrow \infty} f_{\s{ \theta}, k} (\s{x}) f_{\s{ \theta}, k} (\s{x}') \right] = \Sigma(\s{x}, \s{x}').
    \end{align*}
    By using Proposition \ref{prop:srank-covariance} and the independence of the bias vector we evaluate the LHS as
    \begin{align*}
         \Exp\left[ f_{k, \s{\theta}} (\s{x}) f_{k, \s{\theta}} (\s{x}') \right] & = \Exp\left[ 
        \left(\sum_iW_{ki} x_i + b_k \right)\left(\sum_jW_{kj} x_j' + b_k \right)  \right] \\
        & = \frac{s^2 r}{N_1 N_0} \langle \s{x}, \s{x}' \rangle + \sigma_b^2.
    \end{align*}
    
   Lastly, we make some comments about the distribution of $f_{k, \theta}(x)$ as $N_0, N_1 \nearrow \infty$ simultaneously, which will form the base case for induction for the simultaneous limit regime. From the proof above, it is clear that 
    \begin{equation*}
        f_{k, \theta}(x) = \|x\|\frac{ \left( N_1 \sum_a(U_{ka}S_{aa})^2\right)^{1/2} }{\sqrt{N_0N_1}} \tr\left( V. \frac{\sqrt{N_0}(h \otimes x)}{\| h \otimes x\|}\right) + b_k.
    \end{equation*}
    Let $\xi := \frac{\|x\| \left( N_1 \sum_a(U_{ka}S_{aa})^2\right)^{1/2} }{\sqrt{N_0N_1}}$,
    and $\eta := \tr\left( V. \frac{\sqrt{N_0}(h \otimes x)}{\| h \otimes x\|}\right)$. Then, as shown above, we have that $\xi \to \frac{\| x\| (s^2r)^{1/2}}{\sqrt{N_1N_0}}$
    in probability (observe that the natural scaling of $s^2r$ is $\sim N_0N_1$) and $\eta \to N(0, 1)$ in distribution (as an application of Theorem \ref{thm:aristotle}). Since $\xi$ converges to a constant, by a well-known property of convergence in distribution of a product of random variables, we have that the product $\xi\eta$ converges in distribution to $N\left(0, \frac{\|x\|^2 s^2 r}{N_0N_1}\right)$.
    The rest of the proof proceeds similarly.
    
    \end{proof}
     Now, we analyze the application of activations and proceed by induction. At the $l$-th layer we have
    
    \begin{equation}
        \tilde{\s{\alpha}}^l(\s{x}) = \s{W}^l\s{\alpha}^{l - 1}(\s{x}).
    \end{equation}
    
    As in the Lemma, when conditioned on $\s{x}^l, \s{x}'^l$, the joint distribution 
    $(\tilde{\s{\alpha}}^l(\s{x}), \tilde{\s{\alpha}}^l(\s{x}'))$ is $\omega \left( 
    \|\s{\alpha}^{l - 1}(\s{x})\|, \|\s{\alpha}^{l - 1}(\s{x}')\|, \langle \s{\alpha}^{l - 1}(\s{x}), \s{\alpha}^{l - 1}(\s{x}')\rangle\right) $. First, we have by definition that
    
    \begin{equation}
        \langle 
        \tilde{\s{\alpha}}^l(\s{x}), \tilde{\s{\alpha}}^l(\s{x}')\rangle = \sum_{k, s} 
         \alpha_k^{l - 1}(\s{x})\alpha^{l - 1}_k(\s{x}')\sum_i W^l_{ik} W^l_{is}.
    \end{equation}
    From the Lemma, it follows that collection of random variables $\{ Z_i \}_i := \{ W^l_{ik} W^l_{is} \}_i $ is identically distributed and $\cov(Z_i, Z_j) = O(N_l^4)$ using the definition of $\s{W}$ and the Weingarten functions for the orthogonal group for expressions of the type $U_{ia}U_{ib}U_{jc}U_{jd}$. By Chebyshev's inequality
    \begin{align}\label{ineq:law_large_numbers_dependent}
        \Prob \left[ \left| \frac{\sum_i Z_i}{N_l} - \Exp[Z_i] \right| > \epsilon \right]
        &\leq \frac{1}{\epsilon^2} \vari\left(\frac{\sum_i Z_i}{N_l} \right) = \frac{1}{\epsilon^2 N_l^2} \vari \left( \sum_i Z_i \right) \\
        &\leq \frac{1}{\epsilon^2 N_l^2}\sum_i \sum_j \cov(Z_i, Z_j) \nonumber\\
        &\leq \frac{1}{\epsilon^2 N_l^2} \frac{CN_l^2}{N_l^4} \rightarrow 0, \quad N_l \rightarrow \infty.\nonumber
    \end{align}
    Hence, the weak law of large numbers holds and
    \begin{align*}
         \langle 
         \tilde{\s{\alpha}}^l(\s{x}), \tilde{\s{\alpha}}^l(\s{x}')\rangle &\rightarrow \frac{r_l s_l^2}{N_{l - 1}} \langle 
         \s{\alpha}^{l - 1}(\s{x}), \s{\alpha}^{l - 1}(\s{x}')\rangle   = \Exp[\langle 
         \tilde{\s{\alpha}}^l(\s{x}), \tilde{\s{\alpha}}^l(\s{x}')\rangle ],\\
         \| 
         \tilde{\s{\alpha}}^l(\s{x})\|^2 &\rightarrow \frac{r_l s_l^2}{N_{l - 1}} \|
         \s{\alpha}^{l - 1}(\s{x})\|^2 = \Exp[\|
         \tilde{\s{\alpha}}^l(\s{x})\|^2],\\
         \| 
         \tilde{\s{\alpha}}^l(\s{x}')\|^2 &\rightarrow \frac{r_l s_l^2}{N_{l- 1}} \|
         \s{\alpha}^{l - 1}(\s{x}')\|^2 = \Exp[\|
         \tilde{\s{\alpha}}^l(\s{x}')\|^2].
    \end{align*}
    
    In other words, one can read out the parameters of the joint distribution $\omega$ of 
    $(\tilde{\s{\alpha}}^l(\s{x}), \tilde{\s{\alpha}}^l(\s{x}')) $ from the corresponding expectations. To finish the proof, we connect these expectations recursively to the previous layer $l-1$, where this time we take the infinite width limit w.r.t. previous layers. Using the law of large numbers for dependent variables with controlled variance as above (and the Lipschitz property of $\varphi$), in the large width limit we have

    \begin{align*}
         \frac{r_l s_l^2}{N_{l-1}} \|
         \s{\alpha}^{l - 1}(\s{x})\|^2 & \rightarrow r_l s_l^2 \Exp_{z_1} \left[ \varphi(z_1)^2 \right]  =: a_l, \\
         \quad \frac{r_l s_l^2}{N_{l -1}} \| 
         \s{\alpha}^{l - 1}(\s{x}')\|^2 & \rightarrow r_l s_l^2 \Exp_{z_2} \left[ \sigma(z_1) \varphi(z_2) \right] =: b_l,\\
         \frac{r_l s_l^2}{N_{l-1}} \langle 
         \s{\alpha}^{l - 1}(\s{x}), \s{\alpha}^{l - 1}(\s{x}')\rangle & \rightarrow r_l s_l^2 \Exp_{z_1, z_2} \left[ \varphi(z_1) \varphi(z_2) \right] =: c_l,  \\
         \quad (z_1, z_2)&  \sim \omega\left( a_{l-1}, b_{l-1}, c_{l-1} \right).
    \end{align*}
    
    Further, we proceed by induction. This completes the outline of the dynamics and finishes the proof of the lemma.
\end{proof}
Now we record the iterated version of Proposition \ref{lem:large_width_behavior}, which is Theorem \ref{thm:gp-ff} of the main text. The main line of proof is inspired by Proposition 1 of \cite{jacot:neurips:2018} and Theorem D.1 of \cite{Yang_2019} and gives a common argument assimilating both.  
\begin{proof}
The base case of the induction is already covered by Lemma \ref{lem:large_width_behavior}. 
Now, let 
\begin{align*}
   f_{\s{\theta}, i} & = \gamma^l\sum_{j} 
   W^L_{ij}\left(\varphi \left( \tilde{\s{\alpha}}^{L - 1}\right)\right)_j + b_i^L.
\end{align*}
Since the $W^L_{ij}$ and $b_i^L$ are centered, using the independence of the $W_{ij}^L, b^L_i$ from the output of the previous layers, it is clear that $\Exp\left( f_{\s{\theta}, i}\right) = 0$. So, 
\begin{align*}
   \cov\left(f_{\s{\theta}, k}(\s{x}), f_{\theta, k'}(\s{x}')\right)  & = 
   \Exp \left( f_{\s{\theta}, k}(\s{x})f_{\s{\theta}, k'}(\s{x}') \right)\\
    & = (\gamma^l)^2\Exp \left[\left( \sum_j 
    W^L_{kj}  (\varphi  ( \tilde{\s{\alpha}}^{L - 1}(\s{x}; \s{\theta}) ) )_j  \right) \left( \sum_l 
    W^L_{k'l}  (\varphi  ( \tilde{\s{\alpha}}^{L - 1}(\s{x}'; \s{\theta}) ))_l  \right)   \right]  + \delta_{kk'}\sigma_b^2\\
    & = (\gamma^l)^2\sum_{j, l} 
    \Exp\left( W^L_{kj} W^L_{k'l} \right) \Exp
    \left( (\varphi ( \tilde{\s{\alpha}}^{L - 1}(x; \theta)))_j (\varphi (\tilde{\s{\alpha}}^{L - 1}(x'; \s{\theta})))_l\right) + \delta_{kk'}\sigma_b^2 \\
    & = 
    \delta_{kk'}(\gamma^l)^2 \frac{s_L^2r_L}{N_{L - 1}N_L} \sum_j  \Exp \left( (\varphi  ( \tilde{\s{\alpha}}^{L - 1}(x; \s{\theta}) ) )_j (\varphi ( \tilde{\s{\alpha}}^{L - 1}(x'; \s{\theta}) ))_j\right)
    + \delta_{kk'}\sigma_b^2, 
\end{align*}
where we have used the independence of the $L$-th layer from the output $\tilde{\s{\alpha}}^{L - 1}$ of the previous layers. 
By law of large numbers, we have the following convergence in probability 
\begin{align*}
   & \frac{1}{N_{L - 1}} \sum_j  \Exp\left( \varphi \left( \tilde{\s{\alpha}}_j^{L - 1}(\s{x}; \s{\theta})\right) \varphi \left( \tilde{\s{\alpha}}^{L - 1}_j(\s{x}'; \s{\theta})\right)\right) \longrightarrow \Exp_f \left( \varphi(f(\s{x})) \varphi (f(\s{x}')) \right).
\end{align*}
Observe that some care has to be taken in the last step, as the random variables $Z_j := \varphi \left( \tilde{\alpha}_j^{L - 1}(\s{x}; \s{\theta})\right) \varphi \left( \tilde{\alpha}^{L - 1}_j(\s{x}'; \s{\theta})\right)$ are not independent. However, their covariance is $O(1/N^4)$, and one needs to repeat a calculation which is similar to (\ref{ineq:law_large_numbers_dependent}). We skip the repetition. 

Now, 
conditioned on the outputs $\s{\alpha}^{L - 1}$ of the $(L-1)$-th layer, $
f_{\s{\theta}, k}(\s{x})$ are centered Gaussians. In the limit of large width, the covariance, as calculated above, is deterministic and hence independent of $\s{\alpha}^{L - 1}$. So they follow genuine (unconditioned) Gaussian distributions. The claim that the distribution of the vector $\left( 
f_{\s{\theta}, k}(\s{x}_1), \dots, 
f_{\s{\theta}, k}(\s{x}_t)\right)$ is multivariate Gaussian follows similarly as outlined in Lemma \ref{lem:large_width_behavior}. This finishes the proof. 

\end{proof}

\section{Expressivity and curvature analysis}

Here we give an outline of the curvature estimate mentioned in the main text:
\begin{prop}
\label{prop:curvature}
    The feed-forward dynamics of the metric $g^l$ and curvature $\kappa^l$ of the circle are given by 
    \begin{align}
        g^l & = \chi_1 g^{l - 1}, \quad
        \left( \kappa^l\right)^2 = 3\frac{\chi_2}{\chi_1^2} + \frac{1}{\chi_1}\left( \kappa^{l - 1}\right)^2, 
    \end{align}
    where 
    \begin{align*}
        \chi_1   = \frac{(s_t^l)^2 r_t^l}{N_l N_{l-1}}\int \left[\varphi'(\sqrt{q^*}z)\right]^2 Dz, \quad \chi_2  = \frac{(s_t^l)^2 r_t^l}{N_l N_{l-1}}\int \left[\varphi''(\sqrt{q^*}z)\right]^2 Dz.
    \end{align*}
\end{prop}

\begin{proof}
    The proof follows the strategy in Supplementary Sections 2.1-2.3 of \cite{poole:neurips:2016}, only one replaces the standard Gaussian covariance with the one obtained in Proposition~\ref{prop:srank-covariance} and uses Theorem \ref{thm:gp-ff}.
\end{proof}
    
\section{NTK analysis: initialization and training for large widths}\label{sec:7.1_7.2_7.3}

\subsection{Background on NTKs}\label{subsec:ntk-background}

To set the stage, let us suppose that $\b{F}^L : \mathbb{R}^P \to \mathcal{F}, \s{\theta} \mapsto f_\theta$ represents the evaluation function of the DNN having $L$ layers, where the parameter vector $\s{\theta}$ lies in the Euclidean space $\mathbb{R}^P$, and $\mathcal{F} = \{ f_\theta : \s{\theta} \in \mathbb{R}^P \}$ represents the space of output functions $f_{\s{\theta}}$. Usually the training of a DNN will optimize the output $f_{\s{\theta}}$ with respect to a cost functional $\mathcal{C} :\mathcal{F} \to \mathbb{R}$, where the usual gradient descent will be driven by the differential equation $\partial_t \s{\theta} = -\nabla_{\s{\theta}}\mathcal{C}$. However, a nontrivial insight from \cite{jacot:neurips:2018} shows that it is technically easier to lift the equation to the Hilbert space level. Observe that $\mathcal{F}$ becomes a Hilbert space with respect to the inner product $\langle ., .\rangle_{N_0}$ given by $\langle f, g \rangle = \Exp_{\s{x} \sim \b{D}}\left[ f(\s{x})^T g(\s{x})\right]$ for a given data distribution $\b{D}$ on the input space $\mathbb{R}^{N_0}$. Now, using the chain rule, we get
\begin{align*}
    \partial_t \theta_p(t) & = -\partial_{\theta_p}\left( \mathcal{C} \circ \b{F}\right)(\s{\theta}(t)) = -\partial_f \mathcal{C}|_{f_{\s{\theta}(t)}}\partial_{\theta_p} \b{F}^L = - \langle d|_{f_{\s{\theta}(t)}}, \partial_{\theta_p} \b{F}^L_{\s{\theta}(t)} \rangle_{N_0},
\end{align*}
where $\partial_f \mathcal{C}$ denotes the usual Fr\'{e}chet derivative. This implies that the training evolution is given by
\begin{align} \label{eq:grad-descent}
    \partial_t f_{\s{\theta}(t)} & = \sum_p \partial_t\theta_p(t) \partial_{\theta_p} f = -\nabla_{\s{\Theta}^L} \mathcal{C}|_{f_{\s{\theta}(t)}},
\end{align}
where the right hand side denotes the kernel gradient with respect to the NTK $\s{\Theta}^L(\s{\theta})$, which in turn is defined by:
\begin{equation}\label{def:NTK}
    \s{\Theta}^L(\s{\theta}) = \sum_{p = 1}^P \partial_{\theta_p}\b{F}^L(\s{\theta}) \otimes \partial_{\theta_p}\b{F}^L(\s{\theta}).
\end{equation}
For a further detailed discussion of the kernel gradient and the NTK, we refer the reader to Section 3 of \cite{jacot:neurips:2018} and \cite{Hanin2020Finite}.

Now we give a proof of Proposition \ref{thm:Thm_1_Jacot}. To keep the discussion technically less demanding, we give a detailed proof for the sequential case, and indicate how a proof can be given for the simultaneous limit regime, following \cite{Yang_2019}.
   
\begin{proof}[Proof of Proposition \ref{thm:Thm_1_Jacot}]
    Let us first tackle the base case of the induction. Recall that the tangent kernel $\s{\Theta}^1$ is defined by
    \begin{align*}
        \s{\Theta}^1_{kk'}(\s{x}, \s{x}') & = \sum_{p} \partial_{\theta_p} \tilde{ \alpha}^1_k(x; \s{\theta}) \partial_{\theta_p}\tilde{ \alpha}^1_{k'}(\s{x}'; \s{\theta})\\
        & = 
        \sum_p \partial_{\theta_p} \left( \gamma^1\left(\s{W}^1\s{x}\right)_k + \s{b}_k\right) \partial_{\theta_p}\left( \gamma^1\left( \s{W}^1\s{x}'\right)_{k'} + \s{b}_{k'}\right).
    \end{align*}
    Now, $(\s{W}^1\s{x})_k = \sum_i \gamma^1 W^1_{ki}x_i$, which gives that $\partial_{\theta_{mn}} (\s{W}^1\s{x})_k = \sum_i \gamma^1\delta_{mk}\delta_{ni}x_i$. So, we have 
    \begin{align*}
        \s{\Theta}^1_{kk'}(\s{x}, \s{x}') & = 
        \left(\gamma^1\right)^2\sum_{m, n}\sum_{i, j} \delta_{mk}\delta_{ni}\delta_{mk'}\delta_{nj}x_ix'_j + \sum_l \delta_{lk}\delta_{lk'}\\
        & = 
        \left(\gamma^1\right)^2\delta_{kk'}\langle \s{x}, \s{x}'\rangle + \delta_{kk'}.
    \end{align*}
    Now we go for the induction. It is already known that 
    in the large width regime, the preactivations $\tilde{\alpha}^{L - 1}_i$ are centered Gaussian processes. Assume that the NTK $\s{\Theta}_{kk'}^L(\s{x}, \s{x}')$ of the smaller network converges in probability:
    \begin{equation*}
        \sum_p \partial_{\theta_p} \tilde{\alpha}^{L - 1}_k(\s{x}; \s{\theta}) \partial_{\theta_p} \tilde{\alpha}_{k'}^{L - 1}(\s{x}'; \s{\theta}) \longrightarrow \s{\Theta}_\infty^{L - 1}(\s{x}, \s{x}') \delta_{kk'}.
    \end{equation*}
    The new neural net has the old parameters $\theta_p$, and the added parameters on the $L$-th layer, namely, $W^L_{ij}$ and $b^L_k$. We first look at the contribution towards the kernel from the parameters $\theta_p$. Since the new layer is not dependent on the old parameters, we have, 
    \begin{equation*}
        \partial_{\theta_p} f_{\s{\theta}, k}(\s{x}) = 
        \sum_i \gamma^L W^L_{ki} \partial_{\theta_p} \tilde{\alpha}^{L - 1}_i\Dot{\varphi}\left(\tilde{\alpha}^{L-1}_i(\s{x}; \s{\theta})\right).
    \end{equation*}
    We first comment on the case of sequential limits. We have that 
    \begin{align*}
        \sum_p \partial_{\theta_p} f_{\s{\theta}, k}(\s{x}) \partial_{\theta_p} f_{\s{\theta}, k'}(\s{x}') & = 
        \sum_{p, i,j} \left(\gamma^L\right)^2 W^L_{ki} W^L_{k'j}\partial_{\theta_p}\tilde{\alpha}^{L - 1}_i(\s{x}; \s{\theta}) \partial_{\theta_p} \tilde{\alpha}^{L - 1}_j(\s{x}'; \s{\theta})  \Dot{\varphi}\left(\tilde{\alpha}^{L - 1}_i(x; \theta)\right)\Dot{\varphi}\left(\tilde{\alpha}^{L - 1}_j(\s{x}'; \s{\theta})\right) \\
        & = 
        \left(\gamma^L\right)^2\sum_{i, j} W^L_{ki} W^L_{k'j}\s{\Theta}^{L - 1}_{ij}(\s{x}, \s{x}')\Dot{\varphi}\left(\tilde{\alpha}^{L - 1}_i(\s{x}; \s{\theta})\right)\Dot{\varphi}\left(\tilde{\alpha}^{L - 1}_j(\s{x}'; \s{\theta})\right) \\
        & \longrightarrow \left(\gamma^L\right)^2\sum_i \s{\Theta}^{L - 1}_\infty (\s{x}, \s{x}') W^L_{ki}W^L_{k'i}\Dot{\varphi}\left(\tilde{\alpha}^{L - 1}_i(\s{x}; \s{\theta})\right)\Dot{\varphi}\left(\tilde{\alpha}^{L - 1}_i(\s{x}'; \s{\theta})\right)\\
        & \longrightarrow \left(\gamma^L\right)^2\s{\Theta}_\infty^{L - 1}(\s{x}, \s{x}') \delta_{kk'}\frac{r_Ls_L^2}{
        N_L} \Dot{\varphi}^{L}(\s{x}, \s{x}').
    \end{align*}
    where in the second last step we have used the induction hypothesis, and in the last step we have used the weak law of large numbers, as in (\ref{ineq:law_large_numbers_dependent}). 
    The rest of the proof (looking at the contribution of the new parameters $W^L_{ij}, b^L_k$) follows along similar lines as in the base case above.
    
    For the case of simultaneous limits, one can verbatim follow the induction scheme in the proof of Theorem D.1 of \cite{Yang_2019}, inserting the outcome of Proposition \ref{prop:srank-covariance} of the main text at the last step. Since there are no new ideas in our case, we skip writing out the details. However, we make two important observations about the proof of Theorem D.1 of \cite{Yang_2019} which illustrate the essential points of difference with the sequential proof. Firstly, the induction step in  Theorem D.1 starts by expressing the NTK as
    \begin{equation}
        \s{\Theta}^{L}_{kk'}(\s{x}, \s{x}') = \left(\frac{1}{N_l}\sum_{i = 1}^{N_l} \frac{\partial f_{\theta, k}}{\partial \tilde{\alpha}^l_i}(x)\frac{\partial f_{\theta, k'}}{\partial \tilde{\alpha}^l_i}(x')\right)\left( \frac{1}{N_{l - 1}} \sum_{j = 1}^{N_{l - 1}} \alpha_j^{l - 1}(x) \alpha_j^{l - 1}(x') \right),
    \end{equation}
    $l = 1, \dots, L$. This leads to a somewhat different induction scheme (see pp. 25 of \cite{Yang_2019}) than the one above. Secondly, the main technical ingredient introduced in \cite{Yang_2019} that makes the induction argument go through is recorded in the significantly involved Theorem 5.1 which gives the stronger almost sure convergence (instead of convergence in probability) at the last step of the induction. This theorem allows one to justify the gradient independence assumption. Observe that the last step in the sequential induction argument tacitly makes an assumption about the gradient independence. This point has also been explicitly mentioned in \cite{Yang_2019}.
\end{proof}

\begin{rem}[Remarks on the proof of Proposition \ref{thm:large_width_training}]
The proof is long, but virtually similar to the proof of Theorem 2 of \cite{jacot:neurips:2018} (observe that we have stated this result only in the sequential limit regime), so we refer the reader for details therein. A technical requirement for the proof of Proposition \ref{thm:large_width_training} is that one needs to be careful with the normalization $\gamma^l$ in the update rule (\ref{eq:Poole-ff-dynamics}) of the main text. For Proposition  \ref{thm:large_width_training} to hold, one has to choose the normalizations $\gamma^l$ such that 
\begin{equation*}
    \max \{ \gamma^l : 1 \leq l \leq L - 1 \} \to 0 \text{  as } N_1, \dots, N_{L - 2} \to \infty \text{  sequentially.  }
\end{equation*}
This is clear on careful scanning of the proof of Lemma 1 of \cite{jacot:neurips:2018}. We point out that this requirement is not essential for Theorem \ref{thm:gp-ff} and Proposition \ref{thm:Thm_1_Jacot} to work, where we can choose $\gamma^l = 1, \; l = 1, \dots, L - 1$. 
\end{rem}

\begin{rem}
    For overparametrized DNNs, it is observed in practice that gradient descent is largely restricted to a low dimensional linear subspace of the parameter space, which essentially remains fixed throughout the training process \cite{Gur-Ari_2018, Fort_2019}. Such a phenomenon is independent of the initialization, provided one is looking at standard SGD based learning. However, stable rank constrained gradient based learning will with high probability take the learning gradient out of its subspace.
\end{rem}

\subsection{Eigenspectrum of the NTK}\label{subsec:eigenspectrum}
Now we indicate a proof for Proposition \ref{prop:eigenspectrum_NTK} of the main text. 
\begin{proof}
The proof follows from equation (\ref{eq:recursive_NTK}) of the main text. For the sake of simplicity, let us ignore the biases.  Since  $s^2_lr_l \leq C\frac{N_l}{(\gamma^l)^2}$, the expression for the NTK (\ref{eq:recursive_NTK}) of the main text shows that the entries in the kernel matrix are to the order $\sim C$. We know that the eigenvalues are continuous functions of the entries in the kernel matrix (since the roots of a polynomial are continuous functions of the coefficients). This shows that the eigenspectrum of the NTK is smaller in the constrained rank regime and go to $0$ as $N_l \to \infty$. 
\end{proof}

Even with $\gamma^l = \frac{1}{\sqrt{N_l}}$ and the natural assumption $N_{l - 1} \sim N_l$ asymptotically, observe that to get a non-trivial limit in (\ref{eq:recursive_NTK}), we need $s_l^2r_l$ to be $O(N_l^2)$. The main case of interest is then when $s_l^2r_l = O(N_l^2)$, but not $o(N_l^2)$ (because in the latter case the limit is just $0$). This is precisely the case addressed by Proposition \ref{prop:eigenspectrum_NTK} of the main text.

\section{Further empirical results}
\label{sec:further-emp-results}

\subsection{Training speeds}
\label{subsec:training-speeds}

\begin{figure}[h]\label{fig:training_speed}
    \centering
    \includegraphics[width=0.6\linewidth]{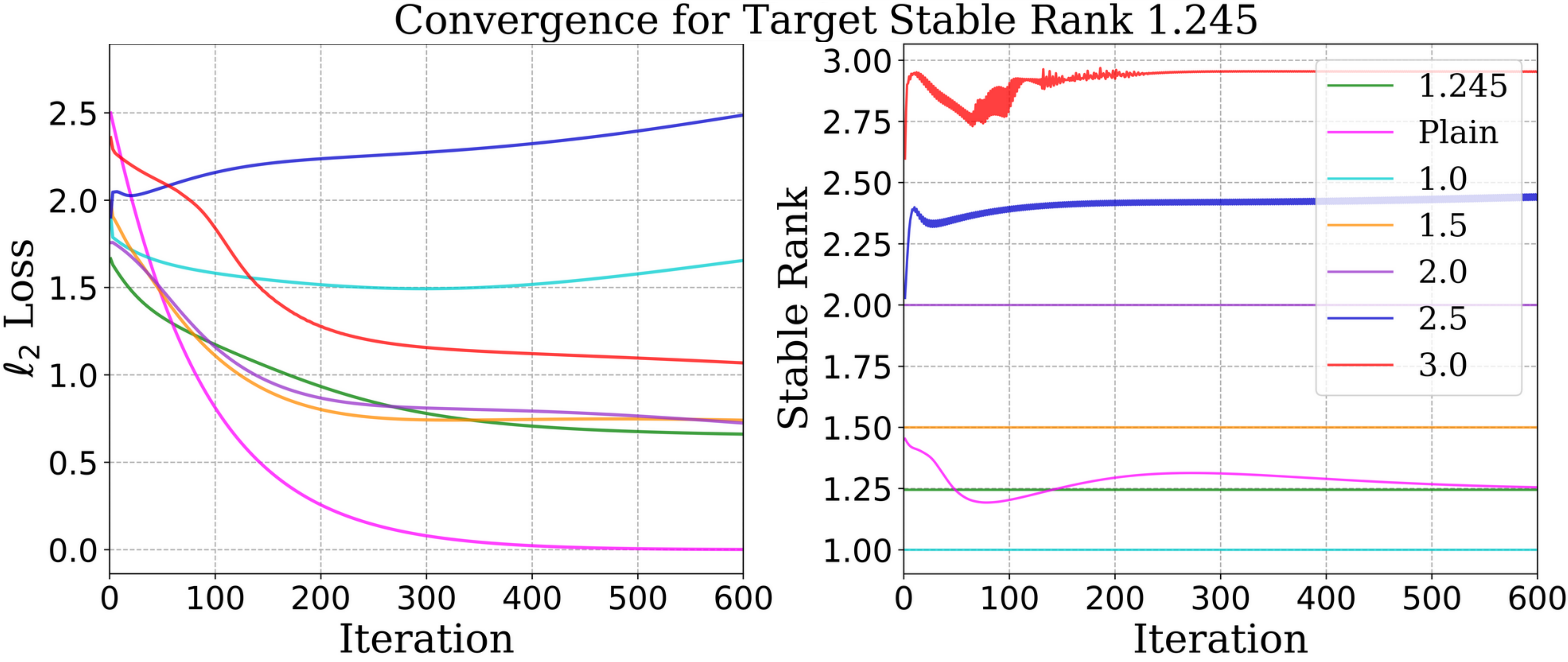}
    \caption{Depiction of training speeds with/without stable rank normalization for a least-squares convex problem. Training behavior of a 3 dimensional linear classifier with various choices of weight matrix stable rank. Enforcing even the target stable rank of the optimal unique solution to a convex problem does not guarantee faster convergence speeds. Natural reasons therefor are the loss landscape and topology.}
    \label{fig:conv-speed}
\end{figure}

Here we elaborate on the toy-example discussed in the main text. 
Suppose we are given 3-sample data in the form of 3-dimensional observation vectors forming a $3\times 3$ diagonal matrix $X$ and suppose to each of the 3 samples a target 3-dimensional label vector is prescribed which also form a $3 \times 3$ diagonal matrix $Y$. Let us regress $(X, Y)$ by fitting a linear classifier described by a $3 \times 3$ diagonal matrix $W$. Of course, the regressor $W^*$ is given explicitly by:
\begin{equation}
    W^* = X^{-1} Y.
\end{equation}
In particular, the optimal solution has a stable rank given by $srank(X^{-1} Y)$. On the other hand, we could start with a given diagonal $W_0$ and converge to $W^*$ via least-squares gradient descent - moreover, the solution traces a line from $W_0$ to $W^*$. We note that in this particular setup, if one applies stable-rank normalization as in \cite{sanyal:iclr:2020} the solution would trace out a curve lying on a sphere (the matrices are diagonal and a fixed stable rank translates to forcing points to lie on a sphere). In particular, the solution curve is necessarily longer than the one for a classical least-squares gradient descent. A numerical description of those observations is given in Fig. \ref{fig:conv-speed}.

\begin{figure}[h]
    \centering
    \includegraphics[width=0.5\linewidth]{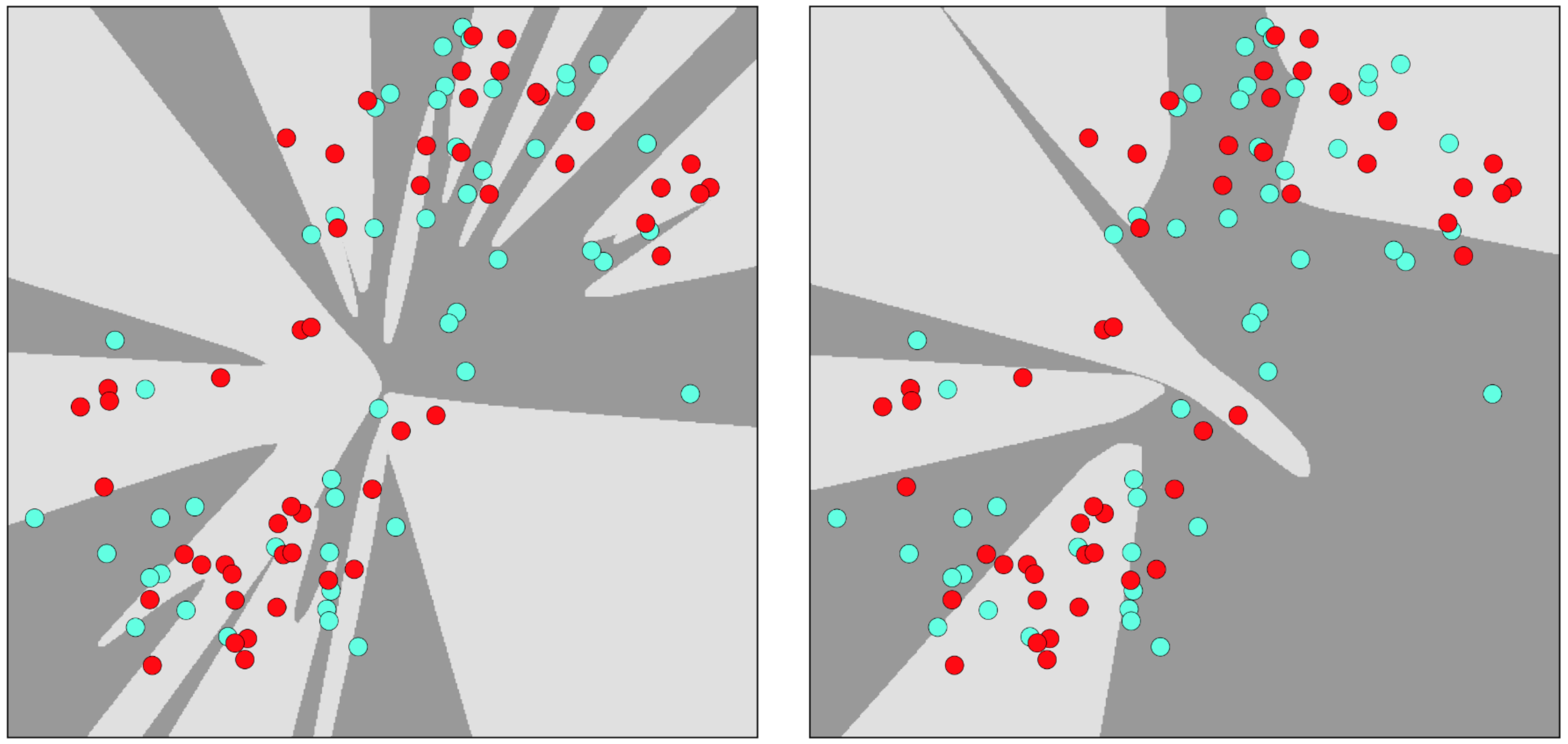} 
    \caption{The effect of stable rank control on a neural network's capability to fit noisy data. Left: a 5-layer MLP is able to fit 2d points with random labeling. Right: the same network architecture with reduced stable rank is unable to achieve sufficient expressivity.}
    \label{fig:decision_boundary_brief}
\end{figure}

\begin{figure}[h]
    \centering
    \includegraphics[width=0.6\linewidth]{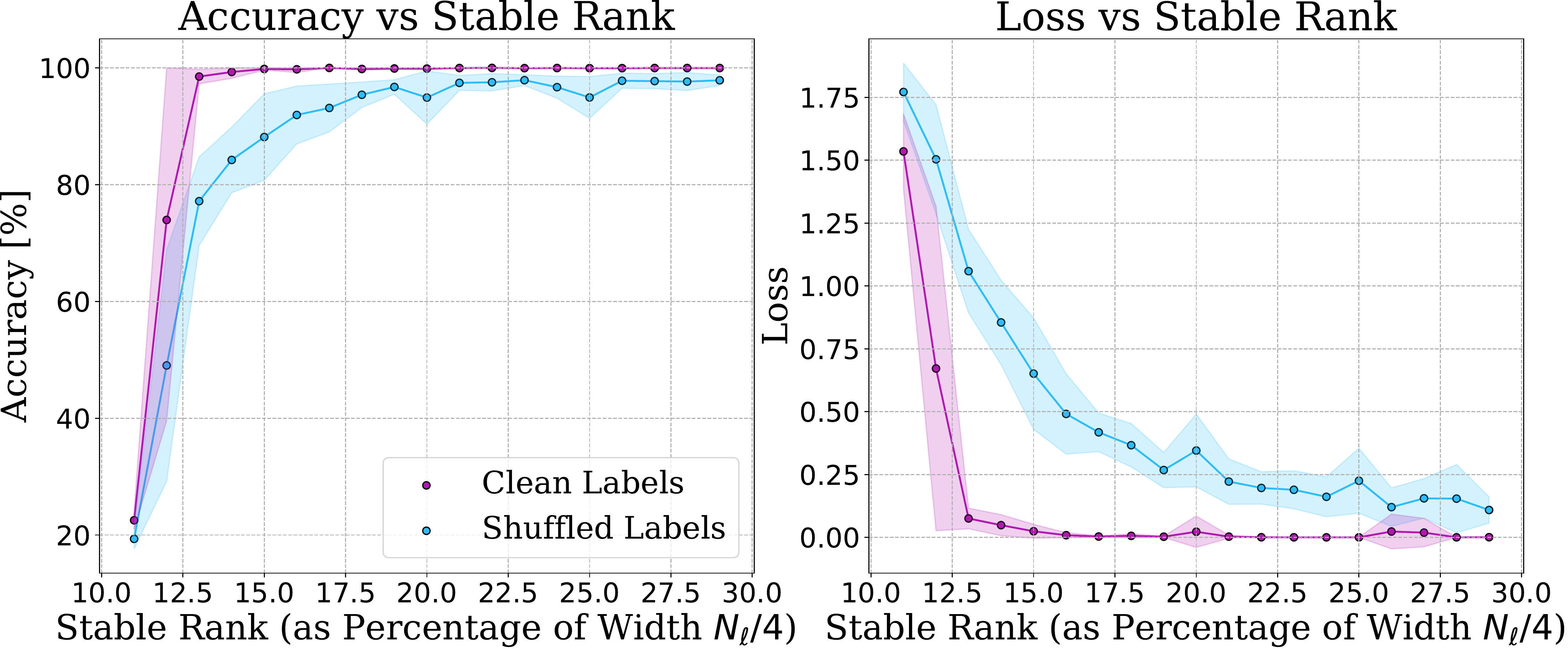}
    \caption{Fitting noisy (shuffled) and clean MNIST data. We use a 2-layer MLP with hidden dimension $h_d = 750$ on a randomly chosen 1000-example subset of the MNIST training set. Sweeping through stable ranks ($x$-axes) we perform 10 trainings per stable rank - the blue (resp. magenta) curves represent the mean accuracy and loss for the shuffled (resp. clean) dataset with transparent areas depicting the variance over the 10 training instances. As expected, the lower stable ranks decrease expressivity and prevent extreme overfitting. Moreover, overfitting the shuffled dataset is clearly a harder problem.}
    \label{fig:shuffled_mnist}
\end{figure}

\textbf{Noise fitting.} Recently, the DNN's striking capability to fit random noise has been put forward \cite{Zhang2017, Maennel2020, Arora2019}. As we have already seen, stable rank control affects the networks geometric expressivity, as intuitively represented by Fig. \ref{fig:decision_boundary_brief}. We now investigate whether stable rank initialization and control during training prevents overfitting of shuffled labels. To this end, we consider the MNIST datset and we compare the DNN's overfitting performance both on the clean (unshuffled) and noisy (shuffled) dataset. The results are presented in Fig. \ref{fig:shuffled_mnist}. On one hand, we observe that in general the DNN needs more time to to overfit the shuffled dataset \cite{Arora2019}; on the other hand, we consistently see that lower stable ranks hinders the overfitting ability capability for both the clean and noisy case. However, this expressivity reduction occurs in a regime where the layers are constrained to $10 - 17 \%$ of their expected stable rank from Proposition \ref{prop:srank-mean-field-nn} of the main text (in the i.i.d. Gaussian case). This suggests a curious "exponential" dependence between stable rank control and noise-fitting capability.

\subsection{Experimental Setup}

The experiments in this work were conducted on a NVIDIA Tesla V100-SXM2-32 GB graphics card. 
We note that while most of the experiments are feasible on less powerful systems, training and test times are greatly reduced when embedded within a GPU computation framework - in particular, we used CUDA 10.2.

\small


\bibliographystyle{plain}
\bibliography{references}

\end{document}